\documentclass[letterpaper]{article}
\usepackage{twocolpaper}

\usepackage[hyphens]{url}
\usepackage{graphicx}
\usepackage{natbib}
\usepackage{caption}
\usepackage{algorithm}
\usepackage{algorithmic}

\usepackage{booktabs}

\usepackage{amsmath}
\usepackage{amssymb}
\usepackage{amsfonts}
\usepackage{mathtools}
\usepackage{xcolor}

\newcommand{\indicator}[1]{\mathbf{1}\{#1\}}

\title{Missingness-Aware Prior-Data Fitted Network for Irregular Clinical Time-Series}
\author{
Hadi Mehdizavareh\textsuperscript{1,*},
Gabriele Santangelo\textsuperscript{2},
Giovanna Nicora\textsuperscript{2},
Pietro Bosoni\textsuperscript{2},
Simon Lebech Cichosz\textsuperscript{1},
Arianna Dagliati\textsuperscript{2},
Arijit Khan\textsuperscript{3,1},
and Riccardo Bellazzi\textsuperscript{2,1}
}

\affiliations{
\textsuperscript{1} Aalborg University, Aalborg, Denmark\\
\texttt{mhme@cs.aau.dk}\\
\textsuperscript{2} University of Pavia, Pavia, Italy\\
\textsuperscript{3} Bowling Green State University, Bowling Green, Ohio, USA
}

\begin{document}

\maketitle

\begin{abstract}
Clinical time series are irregularly observed with laboratory measurements recorded at uneven intervals and observation patterns often reflecting patient state, clinical concern, and site-specific practice. Consequently, missingness is not merely a nuisance factor but an informative signal for prediction. We propose a Missingness-Aware Prior-Data Fitted Network (MA-PFN) for irregular clinical time-series prediction. MA-PFN first learns a diffusion-based prior over joint laboratory value and observation-mask trajectories from source-domain electronic health records, capturing both physiological dynamics and measurement processes. Synthetic trajectories sampled from this prior are then transformed into context-query tasks to train a PFN-style conditional predictor. Once trained, the model performs prediction for new admissions in a single forward pass, requiring neither retraining nor parameter adaptation on target datasets. The framework naturally supports both imputation and forecasting through different context-query constructions. We conduct internal validation on a held-out MIMIC-III test set and external validation on eICU, evaluating both imputation and forecasting and assessing cross-domain generalization under shifts in patient populations, clinical practices, and observation mechanisms. MA-PFN achieves the best performance across all evaluated settings, with Overall/Macro nRMSD ranging from \(0.193/0.186\) for internal imputation to \(0.297/0.297\) for external forecasting, without target-domain adaptation. These results demonstrate that learning missingness-aware task priors enables accurate and transferable clinical time-series prediction under informative missingness.
\end{abstract}


\section{Introduction}
\begin{figure}[t]
\centering
\includegraphics[width=\columnwidth]{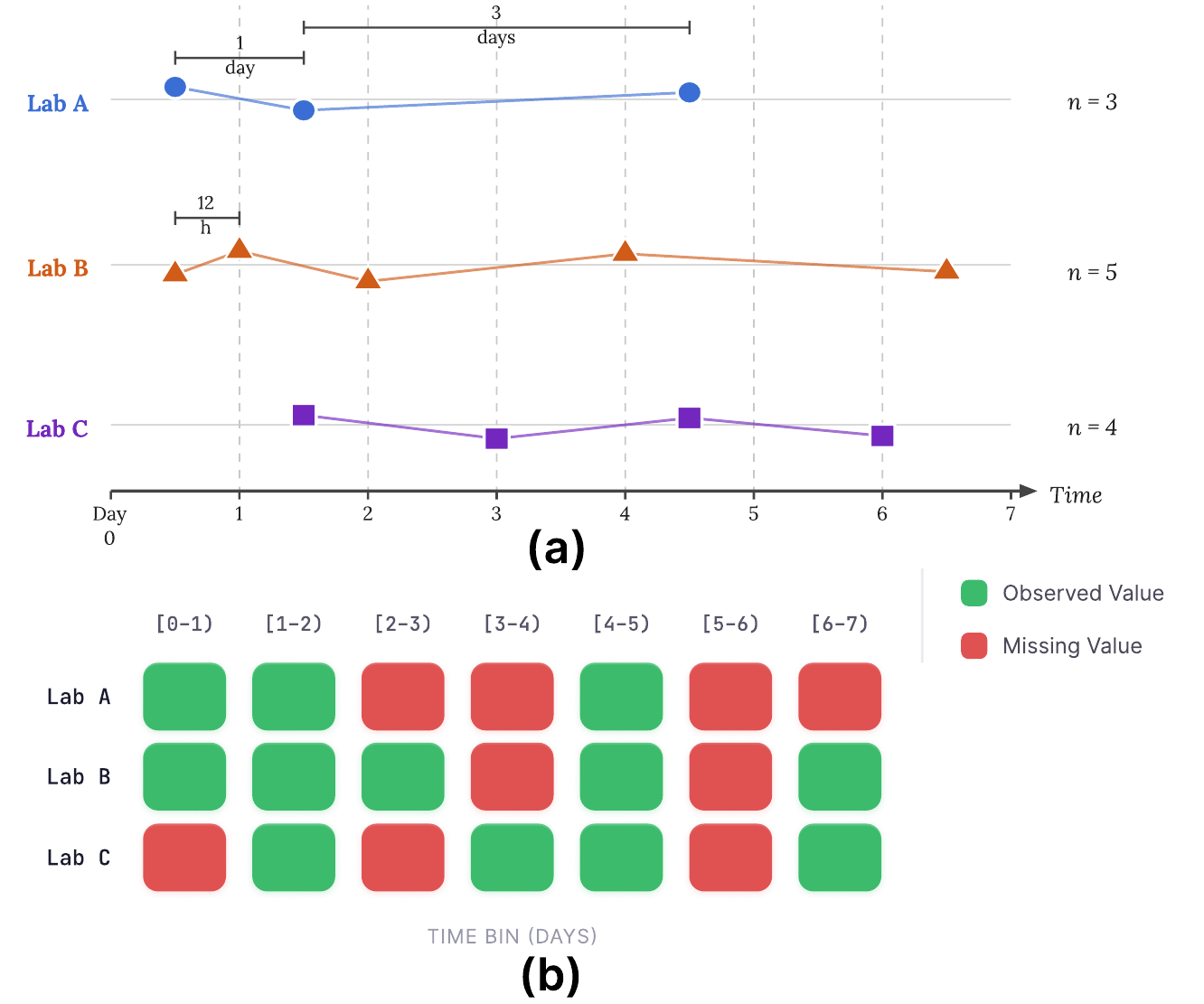}
\caption{Irregular clinical laboratory observations and their value-mask representation. (a) Three laboratory variables measured asynchronously over time. (b) Alignment to a common time grid produces a sparse value-mask matrix, where missing entries reflect unobserved measurements and may contain information about the underlying clinical observation process.}
\label{fig1}
\end{figure}

Electronic health records (EHRs) contain rich longitudinal information about patient state, but they rarely appear as clean, regularly sampled time series \cite{xiao_opportunities_2018, labach_duett_2023}. Laboratory measurements are ordered in response to evolving clinical circumstances rather than according to a fixed data-collection schedule. During the same admission, creatinine may be repeated because of suspected renal deterioration, potassium may be monitored after treatment, and hemoglobin may be measured following concerns about bleeding, while many other variables remain unobserved for extended periods. Consequently, clinical trajectories are characterized by asynchronous measurements across variables and irregular intervals between observations, as illustrated in Fig.~\ref{fig1}(a). When such observations are aligned to a common temporal grid, the result is a sparse value-mask representation like the one shown in Fig.~\ref{fig1}(b). 

Importantly, the resulting missing values are not merely a consequence of incomplete data collection. In clinical practice, the decision to order a laboratory test is itself informative. Repeated testing can reflect concern or active monitoring, whereas the absence of an order may reflect perceived stability, clinical priorities, or site-specific practice \cite{li_imputation_2021, tan_informative_2023}. Effective prediction models must account not only for observed measurements, but also for the observation mechanisms that generated them.

A large body of work has addressed irregular clinical time series through discretization and imputation pipelines, continuous-time latent dynamics, set-based event representations, graph-based architectures, or attention over time and variables \cite{horn_set_2020,tipirneni_self-supervised_2022,kazemi_time2vec_2019,labach_duett_2023,lee_adaptive_2025}. Recent self-supervised approaches further exploit observation patterns, measurement presence, and naturally occurring missingness to learn stronger representations from EHR data \cite{labach_duett_2023,xiang_learning_2025}. These methods substantially improve the encoding of irregular clinical trajectories, yet they largely focus on learning task-specific predictors from finite collections of admissions within a particular dataset. Missingness is typically incorporated as an input feature, reconstruction target, or auxiliary signal; the joint process that produces both laboratory values and their observation patterns is not learned as a reusable prior for prediction. This distinction is important across hospitals, where patient populations, laboratory distributions, and test-ordering policies can all change.

More recently, large-scale EHR foundation models pretrain reusable representations or generative patient models across clinical tasks \cite{steinberg_motor_2024,kraljevic_foresightgenerative_2024,fallahpour_ehrmamba_2025}, but existing approaches largely target coded events, risk outcomes, or general patient trajectories and require downstream adaptation rather than directly supporting missingness-aware continuous laboratory imputation and forecasting. A complementary line of work trains generative models for EHR time series. Mixed diffusion models synthesize continuous measurements alongside discrete variables or missingness indicators \cite{tian_reliable_2024}, but they are mainly assessed as data generators for fidelity, privacy, or augmentation rather than as priors for conditional inference. Prior-data fitted networks (PFNs) instead learn an inference procedure from synthetic tasks and answer new queries in one forward pass \cite{muller_transformers_2022,hollmann_tabpfn_2022}; however, current PFNs rely on generic or hand-crafted priors and fail to capture asynchronous clinical trajectories or informative observation processes.

These developments suggest an opportunity to connect generative modeling and amortized inference. Rather than treating synthetic data generation and clinical prediction as separate objectives, one can learn a clinically grounded prior over laboratory values and observation patterns, and then use that prior to train a reusable inference procedure. Such a strategy would allow prediction models to learn not only from observed admissions but also from a broader distribution of clinically plausible trajectories that capture both physiological evolution and measurement behavior.

\begin{figure*}[t]
    \centering
    \includegraphics[width=\textwidth]{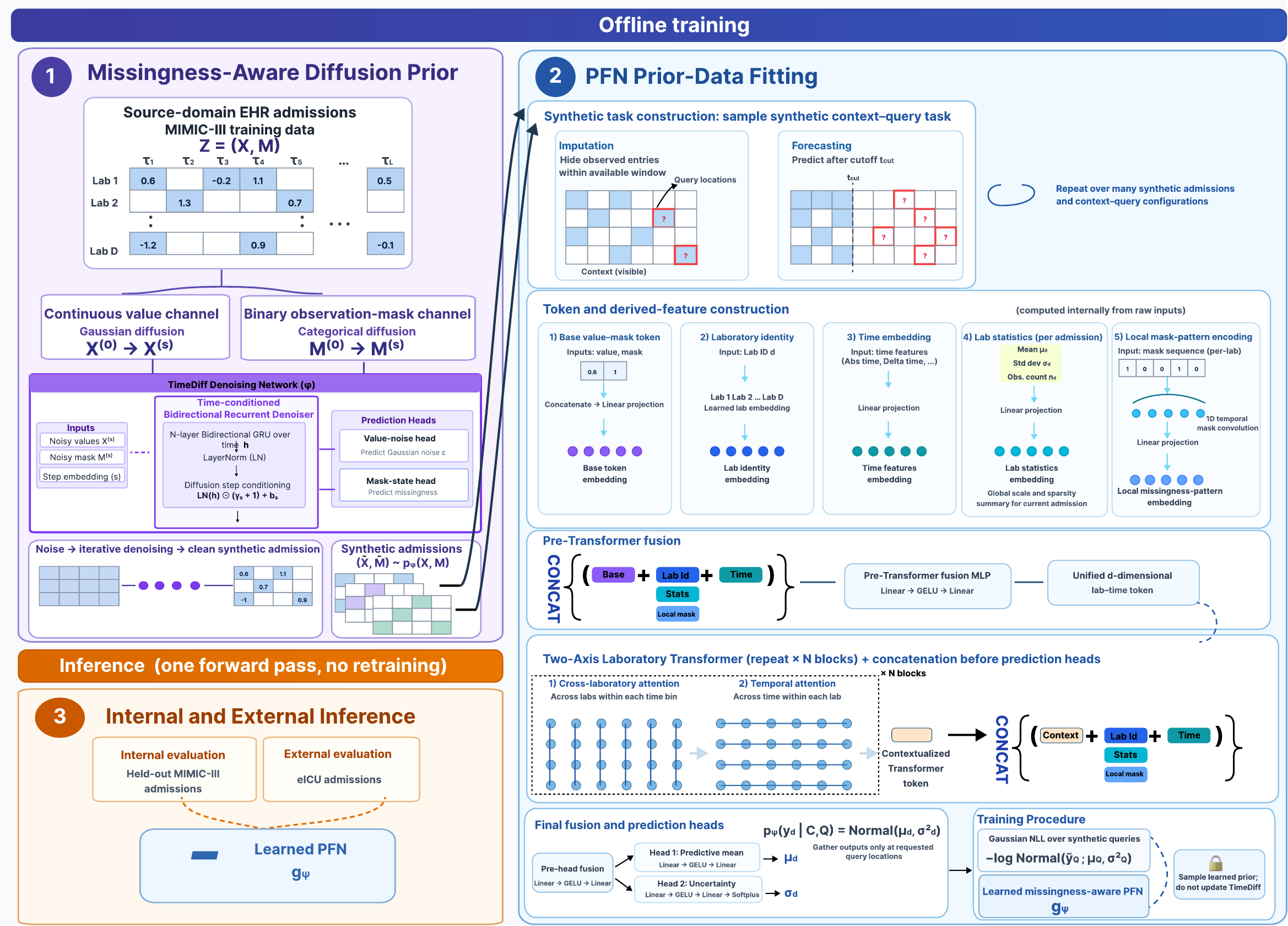}
    \caption{Overview of the proposed Missingness-Aware Prior-Data Fitted Network (MA-PFN) for irregular clinical time-series prediction. 
    }
    \label{fig:overview}
\end{figure*}

We address this gap with the missingness-aware prior-data fitted network (MA-PFN), a framework for irregular clinical time-series prediction. First, a mixed diffusion model learns the source-domain distribution of laboratory value--mask trajectories and generates synthetic admissions that preserve their coupling. Second, these trajectories are converted into context--query tasks used to train a PFN-style conditional predictor. Imputation and forecasting can use the same framework and differ only in which observed values form the context and which locations are queried. After prior fitting, prediction on a real admission requires a single forward pass and no parameter adaptation to the external dataset. To provide an intuitive overview of our framework, Figure~\ref{fig:overview} illustrates the complete MA-PFN pipeline.

MA-PFN differs from prior work in two key ways. First, it treats informative missingness as part of the learned prior rather than only an input feature or auxiliary signal. Second, synthetic trajectories serve not as outputs but as a means to generate large numbers of prediction tasks that train an amortized inference model. Together, these choices enable transferring an inference procedure across domains while preserving clinically meaningful observation patterns. 

\textbf{Contributions.} We (i) formulate irregular clinical imputation and forecasting through a unified context-query prediction framework that explicitly retains informative missingness; (ii) introduce a missingness-aware clinical prior that jointly models laboratory values and observation masks through diffusion-based generation; (iii) show how learned clinical priors can be used to train PFN-style amortized predictors via synthetic task generation; (iv) evaluate both internal and cross-domain generalization on MIMIC-III and eICU, demonstrating the effectiveness of missingness-aware priors for clinical time-series prediction.

\section{Related Work}
\paragraph{Irregular clinical time-series prediction.}
Irregular multivariate time series are characterized by non-uniform sampling intervals and asynchronous observations across variables. Traditional pipelines commonly discretize observations into uniform time bins and impute the resulting missing entries before applying sequence models designed for regularly sampled data \cite{li_imputation_2021,labach_duett_2023,lee_adaptive_2025}. Other methods incorporate irregularity directly: GRU-D uses observation masks and time gaps to decay recurrent states \cite{che_recurrent_2018}; mTAN and SeFT use attention or set functions to represent partially observed events \cite{shukla_multi-time_2021,horn_set_2020}; and latent ordinary differential equations (ODEs), neural controlled differential equations (CDEs), and GRU-ODE-Bayes models describe latent dynamics in continuous time \cite{rubanova_latent_2019,kidger_neural_2020,brouwer_gru-ode-bayes_2019}. More recent graph- and patch-based methods, including Raindrop, GraFITi, tPatchGNN, and APN, capture cross-variable dependencies or adapt local temporal aggregation \cite{zhang_graph-guided_2022,yalavarthi_grafiti_2024, zhang_irregular_2024,liu_rethinking_2025}. Related models such as ATENet learn reference times for irregular-sequence representations, while XTSFormer models the timing and type of future clinical events \cite{lee_adaptive_2025,xiao_xtsformer_2025}. These methods improve task-specific modeling of irregular data, but they are generally trained end-to-end on the finite examples of a particular dataset and do not learn reusable priors.

\paragraph{Informative missingness and self-supervised EHR learning.}
In EHRs, observation patterns can reflect clinical decisions and patient state, motivating models that treat missingness as information rather than only a preprocessing problem \cite{tan_informative_2023}. STraTS represents each observation by its time, variable, and value and uses self-supervised forecasting to learn sparse clinical trajectories \cite{tipirneni_self-supervised_2022}. DuETT places EHRs on a time--event grid, alternates attention across both dimensions, and jointly reconstructs masked values and measurement presence \cite{labach_duett_2023}. AID-MAE distinguishes naturally missing entries from additionally masked observations and learns representations directly from incomplete EHR tables \cite{xiang_learning_2025}; related masked autoencoders focus on tabular or laboratory-value reconstruction \cite{du_remasker_2024,restrepo_representation_2025}. These approaches improve transfer but treat missingness as an input token, mask, or auxiliary reconstruction target. In contrast, MA-PFN learns the joint distribution of values and observation masks as a prior from which complete conditional prediction tasks are sampled.

\paragraph{Diffusion models for imputation and synthetic time series.}
Diffusion models have been applied to imputation \cite{tashiro_csdi_2021, bilos_modeling_2023} and synthetic generation \cite{yoon_time-series_2019,jeon_gt-gan_2022, tian_reliable_2024}. These works evaluate fidelity, privacy, or augmentation, but rarely use diffusion as a task prior for downstream inference. Our approach instead leverages diffusion-generated trajectories to define synthetic context-query tasks, training a predictor that generalizes across domains.

\paragraph{Prior fitted networks (PFNs).}
Bayesian prediction integrates over data-generating hypotheses, but posterior inference is often computationally expensive. PFNs amortize this computation by training a Transformer on synthetic supervised tasks sampled from a prior; after prior fitting, the network directly approximates posterior predictive distributions for new context--query inputs in a single forward pass \cite{muller_transformers_2022}. TabPFN demonstrated this paradigm for small tabular classification using hand-designed priors \cite{hollmann_tabpfn_2022}, while TimePFN extended it to multivariate time-series forecasting \cite{taga_timepfn_2025}. However, existing PFNs rely on generic priors and do not capture asynchronous clinical trajectories or informative missingness. Our work connects missingness-aware EHR generation with PFN training, yielding a reusable inference procedure for imputation and forecasting across datasets.

\paragraph{Summary.}
Prior work has advanced irregular time-series modeling, self-supervised EHR learning, diffusion-based generation, and PFNs. Yet none jointly model laboratory values and observation masks as a clinically grounded prior for conditional prediction. MA-PFN addresses this gap by learning missingness-aware priors and amortizing inference across synthetic tasks, enabling robust generalization under distribution shifts.

\section{Problem Setting}
We study laboratory prediction from irregular EHR trajectories. Each admission is mapped to a fixed clinical time grid $\boldsymbol{\tau}=(\tau_1,\ldots,\tau_L)$, where laboratory variables may be observed at different times. Let $D$ denote the number of laboratory variables. We represent each trajectory as $\mathbf{Z}=(\mathbf{X},\mathbf{M})$, with $\mathbf{X}\in\mathbb{R}^{L\times D}$ containing normalized laboratory values and $\mathbf{M}\in\{0,1\}^{L\times D}$ indicating whether a measurement occurred. We set $M_{\ell d}=1$ when variable $d$ is observed at time $\tau_\ell$ and $M_{\ell d}=0$ otherwise. The mask is treated as part of the clinical trajectory because measurement decisions themselves convey clinical state and monitoring practice.

\paragraph{Context–query formulation.}
A prediction task is defined by a \emph{context} (the visible values and masks)t and and a \emph{query set} (locations to be predicted). Formally, for a query set $\mathcal{Q}\subseteq{1,\ldots,L}\times{1,\ldots,D}$, the context mask $\mathbf{M}^{\mathrm{ctx}}$ is obtained by hiding the queried entries. The conditional prediction problem is then
\begin{equation}
p_\Theta\!\left(\mathbf{y}_{\mathcal{Q}}\mid\mathcal{C},\mathcal{Q}\right),
\label{eq:context_query_problem}
\end{equation}
where $\mathbf{y}_{\mathcal{Q}}=\{X_{\ell d}:(\ell,d)\in\mathcal{Q}\}$ and $\mathcal{C}=\big(\boldsymbol{\tau},\mathbf{X}\odot\mathbf{M}^{\mathrm{ctx}},\mathbf{M}^{\mathrm{ctx}}\big)$. Here $\odot$ denotes elementwise multiplication. Values not visible in the context are zero-filled in the value channel, while their absence is represented by the mask. Thus, the model receives both the observed measurements and the observation pattern.

\paragraph{Imputation vs. Forecasting.} 
Both tasks use the same interface but differ in query selection as shown in the synthetic task construction module at the top of Stage 2 in Figure~\ref{fig:overview}:
\begin{itemize}
    \item \textbf{Imputation:} Queries correspond to values that were observed but deliberately hidden.  
    For example, a known creatinine measurement may be masked and predicted from surrounding context.
    \item \textbf{Forecasting:} A cutoff time \(t_{\text{out}}\) is chosen; the context includes only times before the cutoff, 
    and queries are placed after.  
    For instance, potassium levels after treatment can be predicted from earlier labs.
\end{itemize}

\paragraph{Domain shift.}  
We distinguish a \emph{source domain} \(\mathcal{D}_{\text{src}}\), used to learn priors and train predictors, 
from \emph{target domains} \(\mathcal{D}_{\text{tgt}}\), used for evaluation. 
Target domains may differ in patient populations, measurement frequency, and test-ordering policies 
(e.g., MIMIC-III vs. eICU). 
The goal is to learn a conditional predictor that remains accurate under shifts in both value distributions and observation patterns.

\section{Method}
\subsection{Overview}
MA-PFN consists of two offline stages and one inference stage, illustrated in Figure~\ref{fig:overview}. First, we learn a source-domain generative prior over regularized clinical trajectories. Each trajectory contains a value channel and a mask channel, so the prior models laboratory dynamics together with the observation process. Second, we sample synthetic trajectories from this prior and convert them into conditional prediction tasks by choosing query locations, hiding their values from the context, and training a PFN-style predictor to recover the queried values. At test time, the diffusion model is not used for the target admission. The trained predictor receives the real context and query locations and returns predictions in a single forward pass.

This design separates two roles. The diffusion model learns what plausible source-domain value--mask trajectories look like. The PFN learns how to perform conditional prediction under the task distribution induced by that prior. For both imputation and forecasting, we use the same learned diffusion prior and predictor architecture. The predictor is initially trained on synthetic imputation tasks and subsequently fine-tuned on synthetic forecasting tasks, using a task-specific context–query sampling strategy.

\subsection{Missingness-Aware Diffusion Prior}
The first stage of MA-PFN learns a generative prior over clinical trajectories $\mathbf{Z}=(\mathbf{X},\mathbf{M})$, where $\mathbf{X}$ contains normalized laboratory values and $\mathbf{M}$ contains binary observation mask. Although the data are represented on a fixed grid, irregularity is preserved by the mask channel and by temporal features derived from the observation pattern. The grid provides a common coordinate system; the model does not assume that all variables are measured at all grid times.

The diffusion prior employs a mixed diffusion process: continuous values are corrupted with Gaussian noise, while the binary mask channel undergoes categorical transitions. The denoising network receives the noisy value--mask trajectory, the diffusion step, the grid time, and a recurrence feature indicating how long since each laboratory variable was last observed. The value loss is computed only at observed entries, as unobserved values lack numerical targets. In contrast, the mask loss is evaluated at all lab--time locations, since observation patterns are known for each admission. This dual loss trains the prior to generate both plausible laboratory values and realistic missingness patterns.

After training, synthetic admissions are sampled by running the learned reverse denoising process from noise to a clean value--mask trajectory $(\widetilde{\mathbf{X}},\widetilde{\mathbf{M}})$.  These samples are not used directly as data; instead, they define a \emph{task prior}. By repeatedly converting them into context–query prediction problems, we create a distribution of synthetic tasks that train the PFN predictor in the next stage. This design embeds informative missingness in the prior itself, rather than treating it as an auxiliary feature,  and provides clinically realistic tasks for amortized inference.

\subsection{PFN-Style Conditional Prediction}
\label{sec:PFN_conditional_pred}
Given a sampled trajectory $(\widetilde{\mathbf{X}},\widetilde{\mathbf{M}})$, we construct a conditional prediction task by selecting a query set $\widetilde{\mathcal{Q}}$ from entries whose values are available in the generated trajectory. The selected values become targets, and the corresponding entries are removed from the context mask. The model therefore observes a partially visible value--mask trajectory and must predict the values at the requested query locations. We also perturb visible context values with small noise during training to reduce dependence on exact synthetic values.

Let $\mathcal{T}_{\phi}$ denote the task distribution induced by diffusion sampling, query selection, context construction, and input perturbation. The PFN-style predictor $g_{\psi}$ outputs a mean and variance for each query location. We train it by minimizing the Gaussian negative log-likelihood over synthetic tasks:
\begin{equation}
\begin{split}
\mathcal{L}_{\mathrm{PFN}}(\psi) &= \mathbb{E}_{(\widetilde{\mathcal{C}},\widetilde{\mathcal{Q}},\widetilde{\mathbf{y}}) \sim \mathcal{T}_{\phi}} \Bigg[ \\
&\quad \frac{1}{|\widetilde{\mathcal{Q}}|} \sum_{r\in\widetilde{\mathcal{Q}}} -\log \mathcal{N} \left( \widetilde{y}_{r}; \mu_{\psi,r}, \sigma_{\psi,r}^{2} \right) \Bigg],
\end{split}
\label{eq:pfn_training_loss}
\end{equation}
where $(\boldsymbol{\mu}_{\psi},\boldsymbol{\sigma^2}_{\psi})=g_{\psi}(\widetilde{\mathcal{C}},\widetilde{\mathcal{Q}})$. This objective trains the network to return point predictions and uncertainty estimates. 

Imputation and forecasting are unified under this framework. For imputation, queries correspond to observed values deliberately hidden during training or evaluation. For forecasting, a cutoff time \(t_{\text{out}}\) is chosen, the context is restricted to times \(t \leq t_{\text{out}}\), and queries are placed after the cutoff. In both cases, prediction requires only a single forward pass of the trained PFN, with no parameter adaptation on target domains (Figure~\ref{fig:overview}).

\subsection{PFN Predictor Architecture}
The conditional predictor $g_{\psi}$ is implemented as a two-axis Transformer over the laboratory grid, designed for the context--query tasks in Section~\ref{sec:PFN_conditional_pred}. For each lab--time location $(\ell,d)$, embeddings include:
\begin{itemize}
    \item \textbf{Base evidence:} visible values and context mask.
    \item \textbf{Laboratory identity:} variable-specific embeddings.
    \item \textbf{Time features:} temporal position and recurrence.
    \item \textbf{Laboratory statistics:} mean, standard deviation, and count from the visible context.
    \item \textbf{Local mask pattern:} temporal convolution capturing nearby monitoring behavior.
\end{itemize}

Base, identity, and time embeddings are summed, then concatenated with statistics and mask-pattern features. A fusion MLP maps these into missingness-aware tokens for the Transformer. Attention alternates across grid axes:
\begin{itemize}
    \item \textbf{Cross-laboratory attention:} captures relations among variables at a fixed time.
    \item \textbf{Temporal attention:} models trends and monitoring patterns within each laboratory.
\end{itemize}

Output heads produce predictive means and variances for all lab--time locations, with predictions gathered at query indices. Detailed diffusion equations, task construction, hyperparameters, and implementation settings are provided in Appendices A--C, with the algorithm in Appendix~C.4 describing the full training pipeline.

\subsection{Inference}
At MA-PFN inference, the diffusion prior is no longer used at test time. 
Prediction for a new admission proceeds entirely through the PFN predictor $g_{\psi}$, 
illustrated in Figure~\ref{fig:overview}. Given a real trajectory, the context $C$ and query set $Q$ are constructed as in Section~\ref{sec:PFN_conditional_pred}. 

\section{Experiments}

\subsection{Experimental Setup}
\paragraph{Datasets and preprocessing.}
We use MIMIC-III \cite{johnson_mimic-iii_2016} as the source domain for model development and internal evaluation, and eICU \cite{pollard_eicu_2018} as an independent external domain. Following the DACMI benchmark \cite{luo_evaluating_2022}, we consider 13 commonly measured laboratory tests and retain its predefined admission-level split of 16,534 eligible MIMIC-III admissions into 8,267 training and 8,267 test admissions. The test split is held out throughout development and is used only for internal imputation evaluation. We apply the same laboratory selection and cohort criteria to eICU, yielding 21,423 admissions for external imputation and 10,039 eligible admissions for external forecasting. Both datasets are discretized into 4-hour intervals, retaining the last measurement when a laboratory is recorded multiple times within a bin. Admissions remain variable in length and are padded only within training batches. Training sequences are capped at six weeks (252 time steps) for computational tractability. Laboratory values are transformed using lab-specific log or Box--Cox transformations and standardized using statistics estimated exclusively from the MIMIC-III training split; the same transformations and statistics are applied unchanged to the MIMIC-III test and eICU datasets. Additional cohort and sequence-processing details are provided in Appendix D.1.

\paragraph{Prediction tasks.}
We evaluate three settings: internal imputation on MIMIC-III, external imputation on eICU, and external forecasting on eICU. For internal imputation, we follow the DACMI protocol, where one originally observed value per laboratory is withheld in each MIMIC-III test admission, yielding 13 ground-truth queries per admission \cite{luo_evaluating_2022}. For external imputation, we reproduce this protocol on eICU by randomly withholding one observed value for each of the 13 laboratories and conditioning on the remaining trajectory. Forecasting uses the first two weeks (84 four-hour bins) as context and the subsequent week (42 bins) as the prediction horizon. The imputation model is fine-tuned for forecasting using trajectories sampled from the same learned diffusion prior: synthetic admissions shorter than two weeks are excluded and those exceeding three weeks are truncated, while all observed values in the third week constitute prediction targets. External forecasting applies the resulting model to eligible eICU admissions without target-domain training or adaptation; all observed laboratory values in the third week are evaluated. Additional details about task construction, training procedure, and hyperparameters are provided in Appendix D.2-D.4.

\paragraph{Baselines.}
We compare MA-PFN against representative baselines from several model families. \emph{Na{\"i}ve baselines} include the per-laboratory nad admission mean (Mean), linear interpolation (LI) for imputation, and last observation carried forward (LOCF) for forecasting; LI estimates a held-out value from neighboring observations on both sides of the query, whereas LOCF uses only the most recent observation available before the forecast time. \emph{Traditional machine learning baseline} is represented by XGBoost \cite{chen_xgboost_2016}. \emph{Irregular time-series models} include GRU-D \cite{che_recurrent_2018} and GraFITi \cite{yalavarthi_grafiti_2024}. \emph{Task-specific models} comprise CSDI \cite{tashiro_csdi_2021} for imputation and APN \cite{liu_rethinking_2025} for forecasting. Finally, as a \emph{pretrained prior-fitted baseline}, we evaluate TabPFN \cite{hollmann_tabpfn_2022} on our regularized clinical grid. Further details on baseline inputs, task-specific adaptations, training, and hyperparameters are provided in Appendix D.5.

\paragraph{Evaluation protocol and metrics.}
Our primary metric is range-normalized root-mean-square deviation (nRMSD), following the DACMI evaluation protocol \cite{luo_evaluating_2022}, where range normalization makes prediction errors comparable across laboratory variables with different scales. We report nRMSD in two complementary forms: \emph{Macro nRMSD}, obtained by computing nRMSD separately for each of the 13 laboratories and averaging the resulting scores with equal weight, and \emph{Overall nRMSD}, computed over the pooled prediction targets across all laboratories. We additionally compute mean absolute error (MAE) and root-mean-square error (RMSE) separately for each laboratory; because their magnitudes depend on the laboratory scale, we do not average them across variables and report the complete per-laboratory results in the appendix. Formal metric definitions are provided in Appendix D.6, while additional per-laboratory results are provided in Appendix E.1.

\subsection{Main Results}
Table~\ref{tab:main_results} summarizes performance across internal imputation on MIMIC-III and external imputation and forecasting on eICU. MA-PFN achieves the best Overall/Macro nRMSD in all three settings. For internal imputation, MA-PFN obtains (0.193/0.186), improving over the strongest learned baseline, GraFITi ((0.209/0.205)). The advantage becomes larger under external evaluation: on eICU imputation, MA-PFN reaches (0.204/0.198), corresponding to (13.9\%/15.0\%) lower error than GraFITi ((0.237/0.233)). For external forecasting, MA-PFN obtains (0.297/0.297), slightly outperforming APN ((0.299/0.298)), despite APN being specifically designed for irregular time-series forecasting. 

\begin{figure}[t]
\centering
\includegraphics[width=\columnwidth]{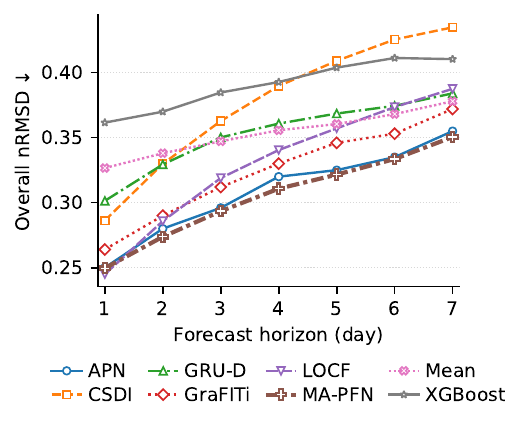}
\caption{External forecasting performance on eICU across the seven-day prediction horizon. For each day, nRMSD is computed separately for each laboratory over its observed query targets and macro-averaged across the 13 laboratories.}
\label{fig:forecast_horizon}
\end{figure}

Simple temporal baselines are notably competitive. LI achieves strong performance on MIMIC-III and eICU imputation, even tying GraFITi in Overall nRMSD on the external dataset. Similarly, LOCF reaches strong performance in forecasting, outperforming Mean, GRU-D, XGBoost, and TabPFN. This behavior is consistent with strong local persistence in laboratory trajectories: when a nearby measurement is available, interpolation or carrying forward the most recent value can remain effective despite substantial irregularity. Their limitations become clearer for longer-range prediction, however, as shown in Fig.~\ref{fig:forecast_horizon}.

TabPFN performs poorly across the three settings, suggesting that a generic tabular prior applied to the regularized grid does not provide sufficient inductive bias for asynchronous temporal dependencies and informative missingness. XGBoost similarly underperforms the time-series models. Unlike typical applications in which gradient-boosted trees benefit from carefully engineered temporal summaries, we intentionally provide XGBoost only the same basic information available to the other methods---time, laboratory values, and observation masks---to avoid giving it additional task-specific information. Laboratory-specific nRMSD, MAE, and RMSE are reported in Appendix~E.1.

\begin{table}[t]
\centering
\small
\setlength{\tabcolsep}{3pt}
\begin{tabular}{@{}lccc@{}}
\toprule
\textbf{Method}
& \textbf{MIMIC-III}
& \textbf{eICU-Imp.}
& \textbf{eICU-Fcst.} \\
\midrule
Mean
& 0.300/0.300 & 0.306/0.306 & 0.349/0.348 \\
LI
& 0.243/0.241 & \underline{0.237/0.236} & \textemdash \\
LOCF
& \textemdash & \textemdash & 0.320/0.320 \\
\midrule
XGBoost
& 0.363/0.359 & 0.406/0.398 & 0.385/0.377 \\
GRU-D 
& 0.261/0.248 & 0.292/0.274 & 0.346/0.341 \\
GraFITi 
& \underline{0.209/0.205} & \underline{0.237/0.233} & 0.312/0.310 \\
CSDI 
& 0.293/0.282 & 0.321/0.319 & 0.366/0.365 \\
APN 
& \textemdash & \textemdash & \underline{0.299/0.298} \\
TabPFN 
& 0.391/0.382 & 0.401/0.401 & 0.387/0.385 \\
\midrule
\textbf{MA-PFN (Ours)}
& \textbf{0.193/0.186} & \textbf{0.204/0.198} & \textbf{0.297/0.297} \\
\bottomrule
\end{tabular}
\caption{Internal and external evaluation measured by Overall/Macro nRMSD. Imp. and Fcst. denote imputation and forecasting, respectively. LI is used only for imputation and LOCF only for forecasting. Best results are shown in \textbf{bold} and second-best results are \underline{underlined}. \textemdash\ indicates a task for which the baseline is not applicable.}
\label{tab:main_results}
\end{table}

Figure~\ref{fig:forecast_horizon} shows that forecasting error generally increases with the prediction horizon across all methods. LOCF is particularly competitive on the first day, highlighting the strength of recent measurements for short-term prediction, but its performance deteriorates rapidly at longer horizons. In contrast, MA-PFN remains the best or among the best methods throughout the horizon, closely tracking or outperforming the forecasting-specific APN while showing a more gradual degradation than the other baselines. Additional results, including training and inference time, per-laboratory forecasting performance, and extended ablation studies, are reported in Appendix E.

\begin{table*}[t]
\centering
\begin{tabular}{@{}lcccccc@{}}
\toprule
& \multicolumn{2}{c}{\textbf{MIMIC-III Imputation}}
& \multicolumn{2}{c}{\textbf{eICU Imputation}}
& \multicolumn{2}{c}{\textbf{eICU Forecasting}} \\
\cmidrule(lr){2-3}
\cmidrule(lr){4-5}
\cmidrule(lr){6-7}
\textbf{Variant}
& \textbf{Overall} $\downarrow$
& \textbf{Macro} $\downarrow$
& \textbf{Overall} $\downarrow$
& \textbf{Macro} $\downarrow$
& \textbf{Overall} $\downarrow$
& \textbf{Macro} $\downarrow$ \\
\midrule
Direct predictor
& 0.200 & 0.198 & 0.235 & 0.230 & 0.325 & 0.325 \\
PFN (decoupled prior)
& 0.238 & 0.237 & 0.242 & 0.240 & 0.309 & 0.305 \\
\textbf{MA-PFN}
& 0.193 & 0.186 & 0.204 & 0.198 & 0.297 & 0.297 \\
\bottomrule
\end{tabular}
\caption{Ablation of prior-data fitting and missingness-aware prior learning. The direct predictor is trained only on real MIMIC-III context--query tasks and does not use prior-data fitting. The decoupled-prior PFN retains synthetic prior-data fitting but learns laboratory values and observation masks independently, removing their joint dependence. MA-PFN uses the full learned joint value--mask prior. Lower nRMSD is better.}
\label{tab:ablation}
\end{table*}

\subsection{Ablation Study}
We isolate the contributions of prior-data fitting and missingness-aware prior learning using two controlled variants of the full MA-PFN. \emph{Direct predictor} uses the identical conditional-prediction architecture but removes the diffusion prior and trains directly on context--query tasks constructed from real MIMIC-III training admissions; for forecasting, it is further fine-tuned on real MIMIC-III forecasting tasks using the same protocol as MA-PFN. This tests whether learning from a reusable synthetic task prior provides benefits beyond the predictor architecture itself. \emph{PFN (decoupled prior)} retains prior-data fitting but replaces the joint prior \(p_{\phi}(\mathbf{X},\mathbf{M})\) with independently learned value and observation processes, \(p_{\phi_X}(\mathbf{X})p_{\phi_M}(\mathbf{M})\). Specifically, the value diffusion model is trained without the mask channel, mask prediction head, mask loss, or mask-derived delta conditioning, while a separate categorical diffusion model learns observation-mask trajectories. Values and masks are then sampled independently and combined before applying the same context--query construction and PFN training procedure. This removes the dependence between what is measured and its value trajectory. Table~\ref{tab:ablation} compares these variants with the full joint value--mask prior. The full MA-PFN achieves the best performance in every setting, but the relative behavior of the two ablations reveals different roles for prior-data fitting and missingness-aware prior learning. On internal MIMIC-III imputation, direct training is already competitive with MA-PFN. Its performance deteriorates much more sharply after transfer to eICU, however. MA-PFN reduces external imputation error relative to direct training by (13.2\%/13.9\%), indicating that the benefit of prior-data fitting is particularly pronounced under domain shift. The same pattern also appears in the forecasting task.

Decoupling laboratory values from observation masks also consistently degrades performance. Compared with PFN (decoupled prior), MA-PFN lowers Overall/Macro nRMSD by (18.9\%/21.5\%) on MIMIC-III imputation, (15.7\%/17.5\%) on external eICU imputation, and (3.9\%/2.6\%) on external forecasting. Although the decoupled model shows little additional MIMIC-to-eICU degradation, this is largely because its internal imputation error is already substantially higher. Taken together, the ablations show that neither the predictor architecture nor synthetic prior-data fitting alone explains the full result: prior fitting improves transfer beyond direct training, while preserving the dependence between laboratory values and observation patterns is critical for achieving strong absolute performance.

\subsection{Limitations and Future Directions}
Our study has the following limitations. MA-PFN is evaluated on 13 numerical laboratory variables from MIMIC-III and eICU, and therefore does not yet establish generalization to the broader heterogeneous EHR. External validation is limited to one independent database, while the learned task prior remains dependent on the MIMIC-III source distribution and on how faithfully the diffusion model reproduces clinically relevant value--missingness relationships. Moreover, discretization into bins can sacrifice exact within-bin timing. Also, the retrospective experiments do not establish prospective clinical utility.

Future work will extend the learned prior to richer EHRs containing a substantially larger set of numerical and categorical variables, including vital signs, medications, diagnoses, procedures, and patient demographics. Learning priors across multiple healthcare systems, relaxing fixed-grid discretization through adaptive or continuous-time representations, and expanding the context--query task family beyond imputation and forecasting are also promising directions. Together, these extensions could move MA-PFN toward a general-purpose EHR foundation model that performs multiple clinical inference tasks without task-specific retraining. An expanded discussion of limitations and future directions is provided in Appendix F.

\section{Conclusion}
We introduced MA-PFN, a missingness-aware prior-data fitted framework for irregular clinical time-series prediction. By learning a joint prior over laboratory values and observation patterns and fitting a conditional predictor on tasks sampled from this prior, MA-PFN provides a unified formulation for imputation and forecasting while requiring no parameter adaptation on the external eICU dataset.  MA-PFN achieves the best Overall/Macro nRMSD across internal MIMIC-III imputation, external eICU imputation, and external eICU forecasting. These findings support learned missingness-aware task priors as a promising approach for transferable prediction from sparse and irregular clinical data.

\bibliography{references}

\clearpage
\appendix

\section{Notation}
\label{app:notation}

Table~\ref{tab:notation} summarizes the notation used in the main paper and appendix. Each example corresponds to one admission after mapping the original irregular measurements to a fixed clinical time grid. The grid is used only as a common coordinate system; irregular observation is represented by the mask and recurrence features.

\begin{table*}[tb]
\centering
\begin{tabular}{ll}
\toprule
Symbol & Meaning \\
\midrule
$i$ & Admission index \\
$L$ & Number of grid time steps \\
$D$ & Number of laboratory variables \\
$\boldsymbol{\tau}=(\tau_1,\ldots,\tau_L)$ & Fixed clinical time grid \\
$\mathbf{X}\in\mathbb{R}^{L\times D}$ & Normalized laboratory-value matrix \\
$\mathbf{M}\in\{0,1\}^{L\times D}$ & Observation mask, where $M_{\ell d}=1$ means observed \\
$\mathbf{Z}=(\mathbf{X},\mathbf{M})$ & Value--mask trajectory for one admission \\
$\mathcal{Q}\subseteq\{1,\ldots,L\}\times\{1,\ldots,D\}$ & Query lab--time locations \\
$\mathbf{M}^{\mathrm{ctx}}$ & Context mask after hiding query values \\
$\mathcal{C}$ & Context supplied to the predictor \\
$\mathbf{y}_{\mathcal{Q}}$ & Target values at query locations \\
$\Delta_{\ell d}$ & Backward recurrence time computed from an observation or context mask \\
$\Delta_{\ell d}^{(s)}$ & Diffusion-step recurrence computed from the current noisy mask state \\
$\phi$ & Parameters of the MA-PFN diffusion prior \\
$\psi$ & Parameters of the PFN predictor in MA-PFN \\
$\mathcal{T}_{\phi}$ & Synthetic task distribution induced by the diffusion prior \\
\bottomrule
\end{tabular}
\caption{Main notation used in the paper and appendix.}
\label{tab:notation}
\end{table*}

\subsection{From Irregular Events to a Value--Mask Grid}
\label{app:irregular_to_grid}

Let the original irregularly observed admission $i$ be
\begin{equation}
    \mathcal{X}_i
    =
    \{(t_{ij},\mathbf{x}_{ij},\mathbf{m}_{ij})\}_{j=1}^{T_i},
    \label{eq:app_original_irregular_series}
\end{equation}
where $t_{ij}$ is the time of the $j$-th measurement event, $\mathbf{x}_{ij}\in\mathbb{R}^{D}$ contains laboratory values, and $\mathbf{m}_{ij}\in\{0,1\}^{D}$ indicates which variables are available at that event. The regularized representation maps these events to a fixed grid $\boldsymbol{\tau}$. For each grid time $\tau_{\ell}$ and laboratory variable $d$, the value channel $X_{\ell d}$ stores the normalized value assigned to that bin when available, and the mask channel $M_{\ell d}$ records whether the value is observed. If more than one value of the same variable falls into the same grid interval, the implementation uses the latest value. The context--query representation used in the main paper can be written explicitly as
\begin{equation}
    \mathcal{C}
    =
    \big(\boldsymbol{\tau}, \mathbf{X}^{\mathrm{ctx}}, \mathbf{M}^{\mathrm{ctx}}, \mathbf{R}^{\mathrm{ctx}}\big),
    \qquad
    \mathbf{X}^{\mathrm{ctx}}
    =
    \mathbf{X}\odot \mathbf{M}^{\mathrm{ctx}},
    \label{eq:app_context_definition}
\end{equation}
where $\mathbf{R}^{\mathrm{ctx}}$ denotes optional recurrence/statistical features derived only from the visible context. For a query set $\mathcal{Q}$, the targets are
\begin{equation}
    \mathbf{y}_{\mathcal{Q}}
    =
    \{X_{\ell d}:(\ell,d)\in\mathcal{Q}\}.
    \label{eq:app_targets_definition}
\end{equation}
The model estimates $p_{\psi}(\mathbf{y}_{\mathcal{Q}}\mid\mathcal{C},\mathcal{Q})$.

\subsection{Imputation and Forecasting Query Construction}
\label{app:query_construction}

For imputation, query locations are sampled from observed entries and hidden from the context:
\begin{equation}
    \mathcal{Q}_{\mathrm{imp}}
    \subseteq
    \{(\ell,d): M_{\ell d}=1\}.
    \label{eq:app_imputation_query_set}
\end{equation}
The corresponding context mask is
\begin{equation}
    M^{\mathrm{ctx}}_{\ell d}
    =
    M_{\ell d}
    \left(1-\indicator{(\ell,d)\in\mathcal{Q}_{\mathrm{imp}}}\right).
    \label{eq:app_context_mask_imputation}
\end{equation}
For forecasting, a cutoff time $t_{\mathrm{cut}}$ is selected. The context is restricted to grid times at or before the cutoff,
\begin{equation}
    M^{\mathrm{ctx}}_{\ell d}
    =
    M_{\ell d}\indicator{\tau_{\ell}\leq t_{\mathrm{cut}}},
    \label{eq:app_context_mask_forecasting}
\end{equation}
and future query locations are selected after the cutoff,
\begin{equation}
    \mathcal{Q}_{\mathrm{for}}
    \subseteq
    \{(\ell,d): \tau_{\ell}>t_{\mathrm{cut}},\ M_{\ell d}=1\}.
    \label{eq:app_forecasting_query_set}
\end{equation}
During evaluation, only locations with known held-out targets are scored.

\section{Detailed Diffusion Prior}
\label{app:diffusion_prior_details}
This section gives the detailed mixed diffusion prior used to generate regularized value--mask trajectories \cite{mehdizavareh_informative_2026}. The prior follows the same principle as mixed continuous--discrete diffusion proposed by TimeDiff \cite{tian_reliable_2024} for EHR time series: continuous laboratory values are handled with Gaussian diffusion, while binary observation masks are handled with categorical diffusion. Each admission is represented as
\begin{equation}
\begin{aligned}
    \mathbf{Z}^{(0)}
    &=
    (\mathbf{X}^{(0)},\mathbf{M}^{(0)}),\\
    \mathbf{X}^{(0)}&\in\mathbb{R}^{L\times D},
    &
    \mathbf{M}^{(0)}&\in\{0,1\}^{L\times D}.
\end{aligned}
\label{eq:app_z0_definition}
\end{equation}
Let $s\in\{1,\ldots,S\}$ denote the diffusion step. We write
\begin{equation}
\begin{aligned}
    \beta_s &\in (0,1),
    & \alpha_s &= 1-\beta_s,\\
    \bar{\alpha}_s &= \prod_{r=1}^{s}\alpha_r,
    & \bar{\alpha}_0 &= 1.
\end{aligned}
\label{eq:app_noise_schedule}
\end{equation}

\subsection{Temporal Conditioning and Recurrence Features}
\label{app:recurrence_features}

The denoising model is conditioned on the fixed grid time and a
diffusion-step-specific backward recurrence feature computed from the
current noisy mask. Let
\begin{equation}
    M_{\ell d}^{(s)}
    =
    \left[
    \overline{\mathbf{M}}_{\ell d}^{(s)}
    \right]_{\mathrm{obs}}
    \in\{0,1\}
    \label{eq:app_noisy_binary_mask}
\end{equation}
denote the observed-state component of the one-hot categorical mask at
diffusion step $s$. For each lab--time location $(\ell,d)$, define
\begin{equation}
\begin{aligned}
    \Delta_{\ell d}^{(s)}
    &=
    \begin{cases}
    0, & \ell=1,\\
    \ell-r_{\ell d}^{\star,(s)},
        & r_{\ell d}^{\star,(s)}\ \mathrm{exists},\\
    \ell-1, & \mathrm{otherwise},
    \end{cases}\\
    r_{\ell d}^{\star,(s)}
    &=
    \max\{r<\ell:M_{rd}^{(s)}=1\}.
\end{aligned}
\label{eq:app_delta_since_last_observed}
\end{equation}
The corresponding step-dependent conditioning array is
\begin{equation}
    \mathbf{H}^{(s)}
    =
    \left\{
    (\tau_{\ell},\Delta_{\ell d}^{(s)}):
    \ell=1,\ldots,L,\ d=1,\ldots,D
    \right\}.
    \label{eq:app_temporal_conditioning}
\end{equation}
Thus, $\mathbf{H}^{(s)}$ is a deterministic function of the fixed grid
and the currently available noisy mask $\overline{\mathbf{M}}^{(s)}$.
It can therefore be computed both during diffusion training and during
every reverse-sampling step, without access to the unknown clean mask.
After each categorical reverse update, the recurrence features are
recomputed from the updated noisy mask before the next denoising step.
At $s=0$, $\Delta_{\ell d}^{(0)}$ reduces to the usual backward
recurrence feature computed from the clean observation mask.

Intuitively, $\Delta_{\ell d}$ tells the model how recently variable $d$ was observed before the current grid time. This feature helps represent monitoring intensity without requiring the model to infer it only from the binary mask at the current location.

\subsection{Gaussian Diffusion for Numerical Values}
\label{app:gaussian_value_diffusion}

For numerical laboratory values, the single-step forward process is
\begin{equation}
    q\!\left(
    X_{\ell d}^{(s)}
    \mid
    X_{\ell d}^{(s-1)}
    \right)
    =
    \mathcal{N}
    \left(
    X_{\ell d}^{(s)};
    \sqrt{\alpha_s}X_{\ell d}^{(s-1)},
    \beta_s
    \right),
    \label{eq:app_value_single_step_forward}
\end{equation}
independently across $(\ell,d)$ conditional on $\mathbf{X}^{(s-1)}$. The cumulative forward distribution has the closed form
\begin{equation}
    q\!\left(
    X_{\ell d}^{(s)}
    \mid
    X_{\ell d}^{(0)}
    \right)
    =
    \mathcal{N}
    \left(
    X_{\ell d}^{(s)};
    \sqrt{\bar{\alpha}_s}X_{\ell d}^{(0)},
    1-\bar{\alpha}_s
    \right),
    \label{eq:app_value_cumulative_forward}
\end{equation}
or
\begin{equation}
    X_{\ell d}^{(s)}
    =
    \sqrt{\bar{\alpha}_s}X_{\ell d}^{(0)}
    +
    \sqrt{1-\bar{\alpha}_s}\epsilon_{\ell d},
    \qquad
    \epsilon_{\ell d}\sim\mathcal{N}(0,1).
    \label{eq:app_value_cumulative_reparameterization}
\end{equation}
The exact Gaussian reverse posterior needed by the variational diffusion objective is
\begin{equation}
\begin{aligned}
    &q\!\left(
    X_{\ell d}^{(s-1)}
    \mid
    X_{\ell d}^{(s)},X_{\ell d}^{(0)}
    \right) \\
    &\qquad =
    \mathcal{N}\!\left(
    X_{\ell d}^{(s-1)};
    \tilde{\mu}_s(X_{\ell d}^{(s)},X_{\ell d}^{(0)}),
    \tilde{\beta}_s
    \right).
\end{aligned}
\label{eq:app_value_exact_reverse_posterior}
\end{equation}
where
\begin{equation}
\begin{split}
    \tilde{\mu}_s(X_{\ell d}^{(s)},X_{\ell d}^{(0)}) 
    &= \frac{\sqrt{\bar{\alpha}_{s-1}}\beta_s}{1-\bar{\alpha}_s}X_{\ell d}^{(0)} \\
    &\quad + \frac{\sqrt{\alpha_s}(1-\bar{\alpha}_{s-1})}{1-\bar{\alpha}_s}X_{\ell d}^{(s)},
\end{split}
\label{eq:app_value_posterior_mean}
\end{equation}
and
\begin{equation}
    \tilde{\beta}_s
    =
    \frac{1-\bar{\alpha}_{s-1}}{1-\bar{\alpha}_s}\beta_s.
    \label{eq:app_value_posterior_variance}
\end{equation}
The learned reverse process for the value channel is parameterized as
\begin{equation}
\begin{split}
    &p_{\phi}\big(
    X_{\ell d}^{(s-1)}
    \mid
    \mathbf{X}^{(s)},
    \overline{\mathbf{M}}^{(s)},
    s,
    \mathbf{H}^{(s)}
    \big) \\
    &\quad =
    \mathcal{N}\big(
    X_{\ell d}^{(s-1)};
    \mu_{\phi,\ell d}^{X},
    \sigma_s^2
    \big),
\end{split}
\label{eq:app_value_learned_reverse}
\end{equation}
where the mean can be expressed through a noise-prediction head
\begin{equation}
    \hat{\epsilon}_{\phi,\ell d}
    =
    f_{\phi,\ell d}^{X}
    \left(
    \mathbf{X}^{(s)},
    \overline{\mathbf{M}}^{(s)},
    s,
    \mathbf{H}^{(s)}
    \right),
    \label{eq:app_value_noise_prediction}
\end{equation}
as
\begin{equation}
    \mu_{\phi,\ell d}^{X}
    =
    \frac{1}{\sqrt{\alpha_s}}
    \left(
    X_{\ell d}^{(s)}
    -
    \frac{\beta_s}{\sqrt{1-\bar{\alpha}_s}}
    \hat{\epsilon}_{\phi,\ell d}
    \right).
    \label{eq:app_value_reverse_mean_noise_param}
\end{equation}
Because true numerical targets are available only when the laboratory
value was observed, the simplified value-channel loss is masked. Using
$\hat{\epsilon}_{\phi,\ell d}
=
f_{\phi,\ell d}^{X}
(\mathbf{X}^{(s)},\overline{\mathbf{M}}^{(s)},s,\mathbf{H}^{(s)})$,
where $\mathbf{H}^{(s)}$ is computed from the same noisy mask
$\overline{\mathbf{M}}^{(s)}$, we write:
\begin{equation}
\begin{aligned}
    \mathcal{L}_{X}(\phi)
    &=
    \mathbb{E}_{\mathbf{X}^{(0)},\boldsymbol{\epsilon},s}
    \left[
    \frac{1}{N_{\mathrm{obs}}}
    \sum_{\ell=1}^{L}\sum_{d=1}^{D}
    M_{\ell d}^{(0)}
    \left(
    \epsilon_{\ell d}-\hat{\epsilon}_{\phi,\ell d}
    \right)^2
    \right],\\
    N_{\mathrm{obs}}
    &=
    \sum_{\ell=1}^{L}\sum_{d=1}^{D}M_{\ell d}^{(0)}.
\end{aligned}
\label{eq:app_value_simple_loss}
\end{equation}

\subsection{Categorical Diffusion for Observation Masks}
\label{app:categorical_mask_diffusion}

For the mask channel, each binary mask entry is represented as a two-class one-hot vector $\overline{\mathbf{M}}_{\ell d}^{(0)}\in\{0,1\}^{2}$. Let $K=2$ be the number of mask classes. The single-step categorical forward transition is
\begin{equation}
    q\big( \overline{\mathbf{M}}_{\ell d}^{(s)} \mid \overline{\mathbf{M}}_{\ell d}^{(s-1)} \big)
    =
    \mathrm{Cat}
    \Big(
    (1-\beta_s)\overline{\mathbf{M}}_{\ell d}^{(s-1)}
    +
    \frac{\beta_s}{K}\mathbf{1}
    \Big).
    \label{eq:app_mask_forward_diffusion}
\end{equation}
Equivalently, with transition matrix
\begin{equation}
    \mathbf{Q}_s
    =
    (1-\beta_s)\mathbf{I}_{K}
    +
    \frac{\beta_s}{K}\mathbf{1}\mathbf{1}^{\top},
    \label{eq:app_mask_transition_matrix}
\end{equation}
the cumulative categorical forward process is
\begin{equation}
\begin{aligned}
    &q\!\left(
    \overline{\mathbf{M}}_{\ell d}^{(s)}
    \mid
    \overline{\mathbf{M}}_{\ell d}^{(0)}
    \right)
    =
    \mathrm{Cat}\!\left(
    \overline{\mathbf{Q}}_s\overline{\mathbf{M}}_{\ell d}^{(0)}
    \right),\\
    &\overline{\mathbf{Q}}_s
    =\mathbf{Q}_s\mathbf{Q}_{s-1}\cdots\mathbf{Q}_1,
    \qquad
    \overline{\mathbf{Q}}_0
    =
    \mathbf{I}_{K}. 
\end{aligned}
\label{eq:app_mask_cumulative_forward}
\end{equation}
The posterior for one categorical reverse step has the standard multinomial-diffusion form
\begin{equation}
    q\!\left(
    \overline{\mathbf{M}}_{\ell d}^{(s-1)}
    \mid
    \overline{\mathbf{M}}_{\ell d}^{(s)},
    \overline{\mathbf{M}}_{\ell d}^{(0)}
    \right)
    =
    \mathrm{Cat}
    \left(
    \frac{\boldsymbol{\varphi}_{\ell d}^{(s)}}
    {\sum_{k=1}^{K}\varphi_{\ell d,k}^{(s)}}
    \right),
    \label{eq:app_mask_reverse_posterior}
\end{equation}
where
\begin{equation}
    \boldsymbol{\varphi}_{\ell d}^{(s)}
    =
    \left(\mathbf{Q}_{s}^{\top}\overline{\mathbf{M}}_{\ell d}^{(s)}\right)
    \odot
    \left(\overline{\mathbf{Q}}_{s-1}\overline{\mathbf{M}}_{\ell d}^{(0)}\right),
    \label{eq:app_mask_reverse_phi}
\end{equation}
and $\odot$ denotes elementwise multiplication. In practice, the denoising network predicts a probability vector for the original mask state,
\begin{equation}
    \widehat{\mathbf{p}}_{\phi,\ell d}^{M}
    =
    f_{\phi,\ell d}^{M}
    \left(
    \mathbf{X}^{(s)},
    \overline{\mathbf{M}}^{(s)},
    s,
    \mathbf{H}^{(s)}
    \right),
    \label{eq:app_mask_x0_prediction}
\end{equation}
and this prediction is used to parameterize the reverse categorical transition. The mask loss is evaluated at all grid locations because the full observation pattern is known:
\begin{equation}
\begin{split}
    \mathcal{L}_{M}(\phi) &= \mathbb{E}_{\mathbf{M}^{(0)},s} \Bigg[ \\
    &\quad - \sum_{\ell=1}^{L}\sum_{d=1}^{D}\sum_{c=1}^{K} \overline{\mathbf{M}}_{\ell d,c}^{(0)} \log \widehat{\mathbf{p}}_{\phi,\ell d,c}^{M} \Bigg].
\end{split}
\label{eq:app_mask_cross_entropy_loss}
\end{equation}
The combined diffusion-prior objective is
\begin{equation}
    \mathcal{L}_{\mathrm{DM}}(\phi)
    =
    \mathcal{L}_{X}(\phi)
    +
    \lambda_M\mathcal{L}_{M}(\phi),
    \label{eq:app_diffusion_combined_loss}
\end{equation}
where $\lambda_M>0$ controls the relative weight of mask reconstruction.

\subsection{Sampling Synthetic Admissions}
\label{app:diffusion_sampling}

After training, synthetic trajectories are obtained by initializing
$(\mathbf{X}^{(S)},\overline{\mathbf{M}}^{(S)})$ from the terminal
noise distributions and repeatedly applying the learned reverse
transitions for $s=S,S-1,\ldots,1$. At each reverse step, the binary
noisy mask $\mathbf{M}^{(s)}$ is obtained from the current categorical
state $\overline{\mathbf{M}}^{(s)}$, and
$\Delta^{(s)}$ and $\mathbf{H}^{(s)}$ are recomputed before evaluating
the denoising network. After sampling
$\overline{\mathbf{M}}^{(s-1)}$, the recurrence features are updated
from this new mask state for the following reverse step. The resulting
sample is
\begin{equation}
\begin{split}
    \widetilde{\mathbf{Z}} &= (\widetilde{\mathbf{X}},\widetilde{\mathbf{M}}) = \mathbf{Z}^{(0)}, \\
    &\quad \mathbf{Z}^{(0)} \sim p_{\phi}(\mathbf{X}^{(0)},\mathbf{M}^{(0)}\mid\boldsymbol{\tau}).
\end{split}
\label{eq:app_synthetic_sample}
\end{equation}
These samples are then used to construct PFN training tasks rather than being reported as released synthetic records.

\section{PFN Objective and Task Construction}
\label{app:pfn_objective}

\subsection{General PFN Formulation}
\label{app:general_pfn}

Prior-data fitted networks amortize Bayesian prediction by training on tasks sampled from a prior \cite{muller_transformers_2022, hollmann_tabpfn_2022}.. Let $\varphi\in\Phi$ denote a latent data-generating mechanism with prior $p(\varphi)$. A context dataset $\mathcal{D}$ and query--target pair $(x,y)$ are generated according to $p(\mathcal{D},x,y\mid\varphi)$. The posterior predictive distribution is
\begin{equation}
    p(y\mid x,\mathcal{D})
    =
    \int_{\Phi}
    p(y\mid x,\mathcal{D},\varphi)
    p(\varphi\mid x,\mathcal{D})
    d\varphi.
    \label{eq:app_pfn_general_ppd}
\end{equation}
A PFN with parameters $\psi$ learns a direct approximation by minimizing the prior-data negative log-likelihood
\begin{equation}
    \mathcal{L}_{\mathrm{PFN}}(\psi)
    =
    \mathbb{E}_{\mathcal{D}\cup\{(x,y)\}\sim p(\mathcal{D})}
    \left[
    -\log p_{\psi}(y\mid x,\mathcal{D})
    \right].
    \label{eq:app_pfn_prior_data_nll}
\end{equation}
Once trained for a prior, the PFN does not fit new parameters for each test dataset; instead, prediction is performed by a forward pass conditioned on the observed context and query inputs. We instantiate this PFN objective in MA-PFN using synthetic clinical tasks induced by the learned diffusion prior.

\subsection{Synthetic Clinical Task Distribution}
\label{app:task_distribution}
Let $\mathcal{T}{\phi}$ denote the synthetic task distribution used to train the clinical PFN. A task is created by sampling a synthetic trajectory from the learned diffusion prior, selecting a set of query locations, hiding the corresponding values from the context, and perturbing the remaining visible context values:
\begin{equation}
(\widetilde{\mathcal{C}},\widetilde{\mathcal{Q}},\widetilde{\mathbf{y}}_{\widetilde{\mathcal{Q}}})
\sim
\mathcal{T}{\phi}.
\end{equation}
This task distribution can be written compactly as
\begin{equation}
\begin{split}
    \mathcal{T}_{\phi} &= p_{\phi}(\widetilde{\mathbf{Z}}^{(0)}\mid\boldsymbol{\tau})
    \pi_{\mathcal{Q}}(\widetilde{\mathcal{Q}}\mid\widetilde{\mathbf{M}}) \\
    &\quad \, p_{\mathrm{obs}}(\widetilde{\mathbf{X}}^{\mathrm{ctx}}
    \mid \widetilde{\mathbf{X}},\widetilde{\mathbf{M}}^{\mathrm{ctx}}).
\end{split}
\label{eq:app_task_distribution}
\end{equation}
Here $p_{\phi}$ is the learned reverse diffusion process over synthetic value--mask trajectories, $\pi_{\mathcal{Q}}$ is the query-selection distribution, and $p_{\mathrm{obs}}$ is the input-noise distribution applied only to visible context values.

Given a sampled trajectory
$(\widetilde{\mathbf{X}},\widetilde{\mathbf{M}})$, the query set is selected directly from observed synthetic entries:
\begin{equation}
\widetilde{\mathcal{Q}}
\subseteq
{(\ell,d):\widetilde{M}_{\ell d}=1}.
\end{equation}
The corresponding targets are
\begin{equation}
\widetilde{\mathbf{y}}_{\widetilde{\mathcal{Q}}}
=
{\widetilde{X}_{\ell d}:(\ell,d)\in\widetilde{\mathcal{Q}}}.
\end{equation}
The context mask is obtained by hiding the queried entries:
\begin{equation}
\widetilde{M}^{\mathrm{ctx}}_{\ell d}
=
\widetilde{M}_{\ell d}
\left(1-\indicator{(\ell,d)\in\widetilde{\mathcal{Q}}}\right).
\end{equation}
The observation-noise component of \(\mathcal{T}_{\phi}\) perturbs only context values (in the normalized training space), not targets. For each laboratory variable \(d\), input-side Gaussian noise is added to observed context values. Specifically, a noise scale is sampled \(\sigma^{\mathrm{obs}}_{d}\sim \mathrm{Unif}(\rho_{\min},\rho_{\max})\), where \(\rho_{\min}\) and \(\rho_{\max}\) are small constants. The context value supplied to the PFN is
\begin{equation}
\begin{split}
    \widetilde{X}^{\mathrm{ctx}}_{\ell d} &= \widetilde{M}^{\mathrm{ctx}}_{\ell d} \left( \widetilde{X}_{\ell d}+\eta_{\ell d} \right), \\
    &\quad \eta_{\ell d}\sim\mathcal{N}(0,(\sigma^{\mathrm{obs}}_{d})^2),
\end{split}
\end{equation}
For locations with \(\widetilde{M}^{\mathrm{ctx}}_{\ell d}=0\), the value channel is set to zero and the absence of a value is represnted by the mask indicators.

This construction gives the PFN the same type of input it receives at test time: a partially visible value--mask context and a set of requested lab--time query locations.

\subsection{Predictive Distribution and Training Loss}
\label{app:pfn_predictive_distribution}
The predictor parameterizes a factorized Gaussian distribution over the queried values:
\begin{equation}
\begin{split}
    p_{\psi}(\widetilde{\mathbf{y}}_{\widetilde{\mathcal{Q}}} \mid \widetilde{\mathcal{C}},\widetilde{\mathcal{Q}})
    &= \prod_{(\ell,d)\in\widetilde{\mathcal{Q}}} \mathcal{N} \big( \widetilde{X}_{\ell d}; \mu_{\psi,\ell d}, \sigma_{\psi,\ell d}^{2} \big), \\
    (\boldsymbol{\mu}_{\psi},\boldsymbol{\sigma}_{\psi}) 
    &= g_{\psi}(\widetilde{\mathcal{C}},\widetilde{\mathcal{Q}}).
\end{split}
\label{eq:app_pfn_gaussian_predictive}
\end{equation}
The training objective is the average Gaussian negative log-likelihood over query locations:
\begin{equation}
\begin{split}
    \mathcal{L}_{\mathrm{imp}}(\psi) 
    &= \mathbb{E}_{\mathcal{T}_{\phi}} \Bigg[ \frac{1}{|\widetilde{\mathcal{Q}}|} \sum_{(\ell,d)\in\widetilde{\mathcal{Q}}} \bigg\{ \frac{1}{2}\log(2\pi\sigma_{\psi,\ell d}^{2}) \\
    &\quad + \frac{(\widetilde{X}_{\ell d}-\mu_{\psi,\ell d})^{2}}{2\sigma_{\psi,\ell d}^{2}} \bigg\} \Bigg].
\end{split}
\label{eq:app_pfn_imputation_nll}
\end{equation}

\subsection{Algorithm}
The complete MA-PFN training procedure is outlined in Algorithm~\ref{alg:two-stage-training}.

\begin{algorithm}[tb]
\caption{Two-stage training of the Missingness-Aware Prior-Data Fitted Network (MA-PFN).}
\label{alg:two-stage-training}
\begin{algorithmic}[1]
\STATE \textbf{Input:} Source admissions $\mathcal{D}_{\mathrm{src}}$; time grid $\boldsymbol{\tau}$; diffusion steps $S$; query sampler $\pi_{\mathcal{Q}}$; diffusion iterations $K_{\mathrm{DM}}$; PFN iterations $K_{\mathrm{PFN}}$
\STATE \textbf{Output:} Trained diffusion prior $p_{\phi}$ and trained predictor $p_{\psi}$
\STATE \textbf{Stage I: Learn the diffusion prior.}
\STATE Initialize diffusion parameters $\phi$.
\FOR{$k=1$ to $K_{\mathrm{DM}}$}
\STATE Sample a mini-batch of admissions from $\mathcal{D}_{\mathrm{src}}$.
\STATE Map each admission to $(\mathbf{X}^{(0)},\mathbf{M}^{(0)})$ on grid $\boldsymbol{\tau}$.
\STATE Sample a diffusion step $s\in{1,\ldots,S}$.
\STATE Corrupt $\mathbf{X}^{(0)}$ with Gaussian diffusion to obtain $\mathbf{X}^{(s)}$.
\STATE Corrupt $\mathbf{M}^{(0)}$ with categorical diffusion to obtain $\overline{\mathbf{M}}^{(s)}$.
\STATE Obtain $\mathbf{M}^{(s)}$ from the observed-state component of $\overline{\mathbf{M}}^{(s)}$ and compute $\Delta^{(s)}$ and $\mathbf{H}^{(s)}$.
\STATE Predict value noise and clean-mask probabilities with $f_{\phi}(\mathbf{X}^{(s)},\overline{\mathbf{M}}^{(s)},s,\mathbf{H}^{(s)})$.
\STATE Update $\phi$ using $\mathcal{L}_{\mathrm{DM}}=\mathcal{L}_{X}+\lambda_{\mathrm{M}}\mathcal{L}_{M}$.
\ENDFOR
\STATE Define $p_{\phi}$ as the learned reverse diffusion process.
\STATE \textbf{Stage II: Train the PFN predictor.}
\STATE Initialize PFN parameters $\psi$.
\FOR{$k=1$ to $K_{\mathrm{PFN}}$}
\STATE Sample $(\widetilde{\mathbf{X}},\widetilde{\mathbf{M}})\sim p_{\phi}(\cdot\mid\boldsymbol{\tau})$ by reverse denoising, recomputing $\Delta^{(s)}$ and $\mathbf{H}^{(s)}$ from the current noisy mask at every reverse step.
\STATE Sample $\widetilde{\mathcal{Q}}\sim\pi_{\mathcal{Q}}(\cdot\mid\widetilde{\mathbf{M}})$ with $\widetilde{\mathcal{Q}}\subseteq{(\ell,d):\widetilde{M}_{\ell d}=1}$.
\STATE Set $\widetilde{\mathbf{y}}_{\widetilde{\mathcal{Q}}}={\widetilde{X}_{\ell d}:(\ell,d)\in\widetilde{\mathcal{Q}}}$.
\STATE Set $\widetilde{M}^{\mathrm{ctx}}_{\ell d}=\widetilde{M}_{\ell d}\big(1-\mathbf{1}{(\ell,d)\in\widetilde{\mathcal{Q}}}\big)$.
\STATE Build $\widetilde{\mathbf{X}}^{\mathrm{ctx}}$ from $(\widetilde{\mathbf{X}},\widetilde{\mathbf{M}}^{\mathrm{ctx}})$.
\STATE Optionally add input-side noise to visible context values.
\STATE Form $\widetilde{\mathcal{C}}=(\boldsymbol{\tau},\widetilde{\mathbf{X}}^{\mathrm{ctx}},\widetilde{\mathbf{M}}^{\mathrm{ctx}},\widetilde{\mathbf{R}}^{\mathrm{ctx}})$.
\STATE Predict $p_{\psi}(\widetilde{\mathbf{y}}_{\widetilde{\mathcal{Q}}}\mid\widetilde{\mathcal{C}},\widetilde{\mathcal{Q}})$.
\STATE Update $\psi$ using the average query negative log-likelihood over $\widetilde{\mathcal{Q}}$.
\ENDFOR
\end{algorithmic}
\end{algorithm}

\section{Additional Experimental Details}
\label{app:additional_experimental_details}

\subsection{Cohort Construction and Preprocessing}
\label{app:cohort_preprocessing}
The MIMIC-III \cite{johnson_mimic-iii_2016} cohort follows the DACMI construction protocol \cite{luo_evaluating_2022}. Admissions are required to contain measurements from each of the 13 selected laboratory variables: chloride, potassium, bicarbonate, sodium, hematocrit, hemoglobin, mean corpuscular volume (MCV), platelet count, white blood cell (WBC) count, red cell distribution width (RDW), blood urea nitrogen (BUN), creatinine, and glucose. Time points at which more than half of these variables are unavailable are removed, and admissions with fewer than 10 remaining time points are excluded. The original DACMI construction additionally excludes three admissions containing a constant value for at least one laboratory variable. These criteria yield 16,534 MIMIC-III admissions, split equally into 8,267 training and 8,267 test admissions. We preserve this split: the training portion is used for development, whereas the test portion is never used for training, model selection, hyperparameter tuning, or estimation of preprocessing parameters.

We apply the corresponding laboratory definitions and cohort-selection procedure to eICU dataset \cite{pollard_eicu_2018} before external evaluation. The resulting external cohorts contain 21,423 admissions for imputation and 10,039 admissions satisfying the forecasting eligibility criteria. The additional forecasting-specific requirements, including the context and prediction horizons, are described with the task definition in Section~\ref{app:task_construction}.

For both databases, laboratory measurements are aligned to consecutive 4-hour bins. If a laboratory is measured multiple times within a bin, its latest measurement is retained. Empty lab--time locations remain missing and are represented by the corresponding observation mask; no temporal imputation is performed during preprocessing. Laboratory-specific log or Box--Cox transformations are applied before z-score standardization. Transformation parameters and normalization statistics are estimated solely from the MIMIC-III training data and subsequently kept fixed for the MIMIC-III test and eICU cohorts.

Admissions are represented at their natural variable lengths rather than divided into fixed windows. During training, sequences are padded to the longest admission within each mini-batch and padding positions are excluded from model computation and loss evaluation. To bound memory requirements, training admissions longer than six weeks are truncated to 252 four-hour bins. Longer admissions are retained during evaluation: they are processed using overlapping 252-step windows with a stride of 14 time steps, and predictions for locations appearing in multiple windows are aggregated. Because such admissions are rare, this procedure is reported here rather than in the main paper.

\subsection{Task Construction}
\label{app:task_construction}
\paragraph{Imputation.}
For both training and evaluation, imputation targets correspond to originally observed measurements that are hidden from the model. Each task contains one query per laboratory variable, for a total of \(D=13\) queries per admission. For synthetic PFN training tasks, one observed location is sampled uniformly at random within each laboratory and withheld from the context. Internal evaluation follows the corresponding DACMI masking protocol on the MIMIC-III test split, with one held-out observed value per laboratory \cite{luo_evaluating_2022}. For eICU external evaluation, we construct the same protocol by uniformly sampling one observed measurement independently for each of the 13 laboratories. Naturally missing entries are never used as evaluation targets because their ground-truth values are unavailable.

\paragraph{Forecasting.}
Forecasting is formulated as prediction beyond a temporal cutoff rather than masking observations within the available context. We use the first 84 four-hour bins (14 days) of an admission as context and the subsequent bins (up to 7 days) as the forecast horizon. All laboratory measurements observed within the forecast horizon are hidden from the predictor and used as query targets; consequently, unlike imputation, the number of targets varies across admissions and laboratories. Admissions shorter than two weeks are excluded, and sequences extending beyond three weeks are truncated at 126 bins. For external eICU evaluation, we additionally require at least \(D=13\) observed laboratory measurements in the forecast horizon in total, without requiring every laboratory to be observed. The same temporal construction is applied to synthetic admissions used for forecasting fine-tuning. Details of how the synthetic prior is generated and how the imputation model is fine-tuned for forecasting are provided in Sec.~\ref{app:prior_training}.

\subsection{Prior Generation and Training Details}
\label{app:prior_training}

\paragraph{Diffusion-prior training.}
The missingness-aware diffusion prior is trained exclusively on the MIMIC-III training split; neither the held-out MIMIC-III test split nor eICU is used for prior learning, model selection, or preprocessing-statistic estimation. Each admission is represented on the 4-hour grid by \(D=13\) laboratory-value channels and their corresponding observation masks. Admissions retain their natural variable lengths and are padded only within mini-batches. To bound memory requirements, training sequences are capped at 252 time steps (six weeks), and only admissions containing at least \(2D=26\) observed laboratory values are retained. The diffusion model jointly learns the continuous laboratory values and discrete observation masks, while naturally missing laboratory values are excluded from the continuous-value loss. Observation recency is additionally provided through delta-time conditioning. We use \(1000\) diffusion steps and the cosine noise schedule adopted by TimeDiff \cite{tian_reliable_2024}. An exponential moving average (EMA) of the diffusion-model parameters is maintained during training and subsequently used for synthetic-sample generation. Detailed optimization settings are reported in Sec.~\ref{app:model_hyperparameters}.

\paragraph{Synthetic prior generation.}
After training the diffusion prior, we generate 256,000 synthetic admissions containing both laboratory values and observation masks. Rather than generating fixed-length sequences, the length of each synthetic admission is sampled from the empirical length distribution of the MIMIC-III training admissions after applying the six-week cap. This preserves variation in admission duration while keeping generated trajectories within the sequence lengths observed during diffusion training. The same minimum-observation criterion used during model development is applied after generation, retaining synthetic admissions with at least \(2D=26\) observed laboratory measurements; approximately \(99\%\) of the generated admissions satisfy this criterion. These retained synthetic trajectories constitute the prior data used for PFN training.

\paragraph{Imputation prior-data fitting.}
The PFN is trained from scratch using the eligible synthetic admissions generated by the diffusion prior. The synthetic trajectories themselves are generated and stored beforehand, whereas context--query tasks are constructed stochastically during PFN training. Specifically, whenever a synthetic admission is used, its imputation query locations are resampled according to the protocol in Sec.~\ref{app:task_construction}; consequently, the same synthetic trajectory can yield different prediction tasks across training iterations. This procedure exposes the predictor to a larger variety of conditional inference problems. No real MIMIC-III test or eICU admissions are used for prior-data fitting. PFN optimization settings are provided in Sec.~\ref{app:model_hyperparameters}.

\paragraph{Forecasting fine-tuning.}
The imputation-trained PFN is subsequently adapted to forecasting by fine-tuning all model parameters on forecasting tasks constructed from the same pool of diffusion-generated synthetic admissions. No additional generative model is trained and no real target-domain data are introduced during this stage. The context, forecast horizon, eligibility criteria, and query construction follow Sec.~\ref{app:task_construction}; only the construction of the conditional prediction task changes from imputation to forecasting. Thus, the forecasting model retains the prior learned through synthetic imputation training while adapting its inference procedure to future-value prediction. Fine-tuning optimization settings are reported in Sec.~\ref{app:model_hyperparameters}.

\subsection{Model Architecture and Hyperparameters}
\label{app:model_hyperparameters}

\paragraph{Diffusion model.}
The diffusion prior follows the TimeDiff bidirectional recurrent denoising architecture \cite{tian_reliable_2024}, with modifications for informative missingness and variable-length laboratory trajectories. The denoiser uses a three-layer bidirectional LSTM with a hidden dimension of 256 and no dropout. At each diffusion step, the noisy laboratory values and masks are concatenated with the normalized delta-time features before entering the recurrent network. Delta time is normalized by the fixed six-week reference length of 252 four-hour bins. Diffusion-step conditioning uses a 64-dimensional sinusoidal embedding that is transformed to match the 256-dimensional recurrent hidden states, followed by the scale-and-shift modulation used in TimeDiff. Layer normalization is applied to the recurrent hidden states before this modulation. We use 1,000 diffusion steps and the cosine beta schedule of TimeDiff. Naturally missing laboratory values are excluded from the continuous diffusion loss, while the continuous-value and discrete-mask objectives are combined using the selected loss weight \(\lambda=10^{-5}\).

The diffusion model is optimized for 200,000 steps using Adam with learning rate \(8\times10^{-5}\) and \((\beta_1,\beta_2)=(0.9,0.99)\). Gradient accumulation over two steps is used, resulting in an effective batch size of 128. An exponential moving average (EMA) of the model parameters is maintained with decay 0.995 and updated every 10 optimization steps; the EMA model is used for synthetic trajectory generation.

\paragraph{PFN predictor.}
The PFN predictor of MA-PFN represents each admission as a lab--time token grid with model dimension \(d=256\). Token representations combine the observed laboratory information with laboratory identity, absolute time, time since the most recent observation, per-laboratory summary statistics, and a local representation of the missingness pattern. Absolute time and delta time are both normalized using the fixed 252-bin reference scale and are jointly projected to the model dimension. Laboratory identity and time representations are added to the value representation. Per-laboratory statistics and local missingness representations are instead concatenated with the resulting token representation and fused using a two-layer MLP with hidden dimension 512 and GELU activation.

Local missingness structure is encoded independently for each laboratory using a one-dimensional convolution over time with kernel size 5. The convolution produces \(d/4=64\) channels, which are linearly projected to the 256-dimensional token space. The same auxiliary representations are fused both before the Transformer and again before the output heads.

The contextual encoder consists of four factorized two-dimensional Transformer blocks. Each block first performs self-attention across the 13 laboratory variables independently at each time point, followed by self-attention across time independently for each laboratory. Both attention operations use four heads, a 512-dimensional feed-forward layer, GELU activation, and zero dropout. We use the standard post-normalization residual structure of PyTorch's \texttt{TransformerEncoderLayer}. To stabilize optimization, the self-attention output projections and the second linear layers of the feed-forward networks are initialized to zero.

The predictive mean and uncertainty are produced by separate two-layer heads, each mapping \(256\rightarrow512\rightarrow1\) with GELU activation. The uncertainty head directly parameterizes the predictive standard deviation, with a Softplus transformation ensuring positivity. For forecasting, an additional per-laboratory head summarizes the hidden representations across the forecast horizon and predicts its mean, standard deviation, and observation fraction; the latter two outputs use Softplus and sigmoid transformations, respectively.

\paragraph{PFN optimization.}
The imputation PFN is trained for 50,000 optimization steps with AdamW, an initial learning rate of \(5\times10^{-4}\), \((\beta_1,\beta_2)=(0.9,0.95)\), weight decay \(0.01\), and batch size 256. The learning rate is linearly warmed up for the first 2,000 steps from \(0.01\) times the target learning rate to \(5\times10^{-4}\), followed by cosine annealing to \(0.1\) times the initial learning rate. Forecasting fine-tuning uses the same optimizer, batch size, learning-rate schedule, and 50,000-step training budget, and updates all model parameters. Table~\ref{tab:ours_hyperparameters} reports the complete selected settings for both the diffusion prior and PFN predictor.

\paragraph{Hyperparameter selection.}
Model development uses only synthetic prediction tasks generated from the learned prior; neither the held-out MIMIC-III test split nor eICU is used for hyperparameter selection. For the PFN, we perform random search over architecture, optimization, and feature-fusion choices. Candidate configurations are assessed from the trajectories of negative log-likelihood (NLL), mean absolute error (MAE), and mean squared error (MSE) on synthetic tasks. The feed-forward dimension is fixed to twice the selected model dimension. For the diffusion prior, the TimeDiff configuration is retained except for the relative weight \(\lambda\) of the discrete missingness objective, which is selected from the candidate values in Table~\ref{tab:ours_hyperparameter_search_compact}. The complete PFN random-search space is reported in the same table.

\paragraph{Computational environment.}
Experiments were conducted on a Linux server equipped with an NVIDIA A10 GPU with 24\,GB of GPU memory, 30 Intel Xeon Ice Lake virtual CPU cores, and 49\,GB of system memory. The software environment used Python~3.10.12, PyTorch~2.2.1, and CUDA~12.1.

\begin{table*}[t]
\centering
\begin{tabular}{llll}
\toprule
\multicolumn{2}{c}{\textbf{Diffusion prior}} &
\multicolumn{2}{c}{\textbf{PFN predictor}} \\
\cmidrule(r){1-2}
\cmidrule(l){3-4}
\textbf{Hyperparameter} & \textbf{Selected value} &
\textbf{Hyperparameter} & \textbf{Selected value} \\
\midrule
Denoiser
    & 3-layer BiLSTM
    & Model dimension \(d\)
    & 256 \\
Hidden dimension
    & 256
    & 2D Transformer blocks
    & 4 \\
Diffusion-step embedding
    & 64
    & Lab-attention heads
    & 4 \\
Diffusion steps
    & 1,000
    & Time-attention heads
    & 4 \\
Beta schedule
    & Cosine
    & Feed-forward dimension
    & 512 \\
Loss weight \(\lambda\)
    & \(10^{-5}\)
    & Dropout
    & 0 \\
Optimizer
    & Adam
    & Mask-convolution kernel
    & 5 \\
Learning rate
    & \(8\times10^{-5}\)
    & Optimizer
    & AdamW \\
Adam betas
    & \((0.9,0.99)\)
    & Learning rate
    & \(5\times10^{-4}\) \\
Effective batch size
    & 128
    & AdamW betas
    & \((0.9,0.95)\) \\
Gradient accumulation
    & 2
    & Weight decay
    & 0.01 \\
Training steps
    & 200,000
    & Batch size
    & 256 \\
EMA decay
    & 0.995
    & Training steps
    & 50,000 \\
EMA update interval
    & 10 steps
    & Warmup steps
    & 2,000 \\
---
    & ---
    & Minimum learning rate
    & \(0.1\times\) initial LR \\
---
    & ---
    & Forecast fine-tuning
    & 50,000 steps \\
\bottomrule
\end{tabular}
\caption{Selected architecture and optimization hyperparameters for the two components of MA-PFN: the missingness-aware diffusion prior and PFN predictor.}
\label{tab:ours_hyperparameters}
\end{table*}

\begin{table*}[t]
\centering
\begin{tabular}{ll}
\toprule
\textbf{Hyperparameter} & \textbf{Search space} \\
\midrule
\multicolumn{2}{l}{\textit{Diffusion prior}} \\
Loss weight \(\lambda\)
    & \(\{10^{-6},10^{-5},10^{-4},10^{-3},10^{-2},10^{-1},1\}\) \\
\midrule
\multicolumn{2}{l}{\textit{PFN predictor}} \\
Training steps
    & \(\{50\mathrm{k},100\mathrm{k},200\mathrm{k}\}\) \\
Model dimension (\(d\))
    & \(\{64,128,256,512\}\) \\
Transformer blocks
    & \(\{4,8,12\}\) \\
Attention heads
    & \(\{4,6,8\}\) \\
Feed-forward dimension
    & \(2d\) \\
Learning rate
    & \(\{10^{-5},10^{-4},5\times10^{-4},10^{-3}\}\) \\
Batch size
    & \(\{32,64,128,256,512\}\) \\
Lab-statistics fusion
    & \(\{\mathrm{add},\mathrm{concat}\}\) \\
Mask-convolution fusion
    & \(\{\mathrm{add},\mathrm{concat}\}\) \\
\bottomrule
\end{tabular}
\caption{Hyperparameter search spaces used for MA-PFN model selection. Random search was used for the PFN predictor, while the diffusion prior retained the TimeDiff configuration except for the missingness-loss weight.}
\label{tab:ours_hyperparameter_search_compact}
\end{table*}

\subsection{Baselines}
\paragraph{Common experimental protocol.}
All baselines received the same laboratory measurements and preprocessing used by MA-PFN. Numerical values were transformed and standardized using parameters estimated exclusively from the MIMIC-III training split. Observation masks and elapsed-time information were derived only from measurements visible in the context. Models operating on a regular grid received the 4-hour value--mask representation described in Appendix~\ref{app:cohort_preprocessing}. For methods designed for irregular observations, charttimes (event timestamps), event types and values were given directly to the methods.

Trainable baselines were developed using only the MIMIC-III training split. When early stopping or model selection was required, an admission-level subset of this training split was used for validation. The predefined MIMIC-III test split and all eICU admissions remained unseen during training and hyperparameter selection. The selected MIMIC-III models were applied unchanged to eICU, with no target-domain fine-tuning, recalibration, preprocessing re-estimation, or hyperparameter selection.

For imputation, one observed value per laboratory was hidden from each eligible admission during training, following the evaluation task construction. For forecasting, models received the first 84 grid positions, corresponding to two weeks, and predicted positions 84--125, corresponding to the subsequent 42-position week. Future laboratory values and future observation masks were never supplied as model inputs. Separate task-specific checkpoints were trained when a method required a different objective or input construction for imputation and forecasting. Table~\ref{tab:baseline_hyperparameters} summarizes the principal settings.

\paragraph{Mean.}
The Mean baseline predicted the source-domain training mean independently for each laboratory and admission. Mean was evaluated for both imputation and forecasting and used no information from the target admission beyond the queried laboratory identity.

\paragraph{Linear interpolation.}
Linear interpolation (LI) was used only for imputation. For a query at time $\tau_q$ and laboratory $d$, let $(\tau_p,x_p)$ and $(\tau_n,x_n)$ denote the nearest observed measurements of that laboratory before and after the query, respectively. When both were available, the prediction was
\begin{equation}
\widehat{x}_{q,d}
=
x_p
+
\frac{\tau_q-\tau_p}{\tau_n-\tau_p}
(x_n-x_p).
\label{eq:baseline_linear_interpolation}
\end{equation}
If observations were available on only one side of the query, the nearest available endpoint was used. If the admission contained no visible value for that laboratory, the MIMIC-III training mean was used. LI was not evaluated for forecasting because interpolation after the cutoff would require observations from the forecast horizon.

\paragraph{Last observation carried forward.}
Last observation carried forward (LOCF) was used only for forecasting. At each future query time, the prediction was the most recent observed value of the same laboratory available at or before the context cutoff. The value was then carried forward throughout the forecast horizon until another context observation was available; no measurements from the forecast week were revealed. If a laboratory had no observation in the two-week context, its MIMIC-III training mean was used. LOCF was not included as an imputation baseline because it is intrinsically one-sided, whereas the imputation setting permits observations on both sides of an internal query.

\paragraph{XGBoost.}
We trained 13 independent XGBoost regressors, one for each laboratory \cite{chen_xgboost_2016}. Each supervised example corresponded to one lab--time query. Its query-conditioned feature vector was constructed entirely from the visible context and contained the query laboratory identity, normalized query time, admission length, context laboratory values, observation indicators, and time-since-observation features. We deliberately did not add domain-specific feature engineering, ensuring that XGBoost received the same underlying information as the sequence models.

For imputation, training queries were sampled from observed entries, with one query per available laboratory in each sampled admission. Admissions of at most 252 positions were processed directly. Longer admissions were evaluated using 252-position overlapping windows with stride 128; predictions from overlapping windows were averaged at the corresponding global lab--time locations.

For forecasting, a separate set of regressors was trained using 126-position sequences containing an 84-position context and a 42-position horizon. Features for every future query were calculated using only the context. In particular, neither the values nor the observation pattern of the third week were included in the feature construction. The models predicted the full future grid, after which evaluation selected the positions at which ground-truth future measurements were available. We used the squared-error regression objective.

\paragraph{GRU-D.}
GRU-D incorporates laboratory-specific temporal decay based on observation masks and time gaps \cite{che_recurrent_2018}. The input at each grid position consisted of normalized laboratory values, binary observation indicators, and elapsed-time variables. Missing numerical entries were initialized to zero in standardized space, while the GRU-D decay mechanism combined the most recently observed value with the empirical MIMIC-III training mean.

We used the bidirectional GRU-D variant for both tasks. In imputation, bidirectional processing allows the prediction to use visible observations before and after an internal query. The model was trained with one randomly hidden observed value per laboratory and an MSE objective. Training sequences were capped at 252 positions and padded only within each mini-batch. Admissions longer than 252 positions were processed by overlapping windows with stride 128, with predictions averaged in overlapping regions.

For forecasting, a separate checkpoint was trained on fixed 126-position sequences. Only the first 84 positions contained real context values and masks. Every position in the 42-step future segment was represented as unobserved, so the bidirectional recurrent computation did not receive the true future value or future measurement pattern. All observed future values were used as regression targets during training and evaluation. The implementation used one recurrent layer, recurrent dropout of 0.3, output-head dropout of 0.5, and MSE loss; the empirical laboratory means were estimated from the MIMIC-III training data and stored with the checkpoint.

\paragraph{GraFITi.}
GraFITi represents an irregular time series as a sparsely observed bipartite graph between measurement events and variables and applies graph-based attention for conditional prediction \cite{yalavarthi_grafiti_2024}. We supplied the model with irregular event tuples derived from the same laboratory values, observation times, and masks used by the other methods. The model received no static patient variables or auxiliary clinical features.

For imputation, each deliberately hidden measurement was represented as a query event located within the temporal support of the admission, while all remaining observations formed the conditioning set. One observed query was sampled per available laboratory during training. For forecasting, the conditioning set contained events from the first two weeks and the query set contained lab--time locations in the following week. GraFITi was trained separately for the two tasks using MSE over the corresponding query values.

We used a latent dimension of 128, four attention heads, and one graph-attention layer. Optimization used AdamW with learning rate $10^{-3}$ and mini-batches of 64 admissions. Training was limited to 200 epochs, with a ReduceLROnPlateau scheduler using a reduction factor of 0.5, scheduler patience of 10 epochs, and minimum learning rate of $10^{-5}$. Early stopping used a patience of 30 epochs.

\paragraph{CSDI.}
CSDI is a conditional score-based diffusion model originally introduced for probabilistic time-series imputation \cite{tashiro_csdi_2021}. It was applied to the 4-hour value--mask grid and conditioned on observed values, observation masks, time embeddings, and laboratory embeddings. The target mask identified values to be generated, whereas naturally missing values that were not selected as evaluation targets contributed neither conditioning values nor regression targets.

For imputation, the target mask contained one randomly selected observed entry per available laboratory. At inference, it contained the fixed DACMI queries for MIMIC-III or the corresponding sampled eICU queries. For forecasting, we adapted the same conditional-generation formulation by treating all observed values in the third week as targets and restricting the conditioning mask to the first 84 positions. Thus, CSDI was evaluated for forecasting without access to either future values or the future observation pattern. The forecasting adaptation retained the standard attention mechanism used for imputation rather than enabling the linear-attention option.

The network used four residual layers, 64 residual channels, eight attention heads, a 128-dimensional diffusion-step embedding, a 128-dimensional time embedding, and a 16-dimensional laboratory embedding. We used 50 diffusion steps with a quadratic noise schedule from $\beta_1=10^{-4}$ to $\beta_{50}=0.5$. Optimization used Adam with learning rate $10^{-3}$. The imputation model used mini-batches of 16 and up to 200 epochs; the forecasting model used mini-batches of 8 and up to 100 epochs. At inference, 50 conditional samples were generated and their sample mean was used as the point prediction reported in the deterministic error metrics.

\paragraph{APN.}
The Adaptive Patch Network (APN) is specifically designed for irregular multivariate time-series forecasting and was therefore evaluated only in the forecasting setting \cite{liu_rethinking_2025}. Irregular context observations from the first two weeks were supplied as time--variable--value events, and APN constructed adaptive temporal patches from these observations. All observed values in the subsequent week were used as prediction targets. Values and observation masks from the forecast horizon were not included in the input.

We adapted the original forecasting interface to 13 laboratory variables, an 84-position context, and a 42-position horizon. The model dimension and time-encoding dimension were both 16, the maximum number of adaptive patches was 50, and dropout was zero. APN was optimized using Adam with learning rate $0.02$, batch size 32, and MSE loss. Training was limited to 200 epochs with early-stopping patience of 10 epochs.

\paragraph{TabPFN.}
For the prior-fitted tabular baseline, we used the official pretrained \texttt{TabPFNRegressor} model distributed through the TabPFN repository%
\footnote{\url{https://github.com/PriorLabs/TabPFN}}%
\cite{hollmann_tabpfn_2022}. We did not train a new TabPFN prior or update the pretrained Transformer weights on MIMIC-III or eICU.

Because TabPFN expects a tabular regression dataset rather than a multivariate sequence, each admission was converted into a cell-wise regression problem. An observed cell $(\ell,d)$ became one in-context training example. Its feature vector contained the time index $\ell$, laboratory index $d$, the partially observed row at time $\ell$, and the partially observed temporal column for laboratory $d$. The target was the normalized value $X_{\ell d}$. Cells selected as imputation or forecasting queries were converted to unlabeled test examples using the same representation. Missing entries remained represented as \texttt{NaN}, allowing the pretrained TabPFN preprocessing to identify unavailable components.

For imputation, observed non-query cells from the complete admission formed the in-context training set, and the deliberately hidden cells formed the test set. For forecasting, only cells from the first 84 positions were included as labeled context examples, while cells in the subsequent 42 positions were supplied as unlabeled queries. Consequently, the pretrained model did not receive future laboratory values or future labels. We used an ensemble size of 10 and disabled the optional additional mask and recurrence feature blocks, because the cell-wise row--column representation already retained the missing entries. TabPFN was fitted independently in context for each admission, without gradient-based parameter updating or target-domain adaptation.

\begin{table*}[tb]
\centering
\small
\setlength{\tabcolsep}{3.5pt}
\begin{tabular}{@{}p{1.55cm}p{1.35cm}p{4.05cm}p{9.05cm}@{}}
\toprule
\textbf{Method} &
\textbf{Tasks} &
\textbf{Representation} &
\textbf{Principal hyperparameters and training settings} \\
\midrule

Mean &
Imp./Fcst. &
Laboratory identity &
Per-laboratory MIMIC-III training mean; no trainable parameters. \\

LI &
Imp. &
Per-laboratory observations and times &
Piecewise-linear interpolation; nearest-endpoint extrapolation; training-mean fallback when no context value exists. \\

LOCF &
Fcst. &
Per-laboratory context history &
Most recent pre-query observation; MIMIC-III training-mean fallback; no trainable parameters. \\

XGBoost &
Imp./Fcst. &
Query-conditioned grid features &
Thirteen independent regressors; squared-error objective; at least two observed training values per laboratory; 252-step imputation windows with stride 128; forecasting context and horizon of 84 and 42 steps, respectively. \\

GRU-D &
Imp./Fcst. &
Values, masks, and forward/backward elapsed times &
Bidirectional GRU-D; one recurrent layer; recurrent dropout 0.3; prediction-head dropout 0.5; empirical-mean decay; MSE loss; 252-step imputation windows with stride 128; forecasting context and horizon of 84 and 42 steps, respectively. \\

GraFITi &
Imp./Fcst. &
Irregular time--variable--value events &
Latent dimension 128; one graph-attention layer; four attention heads; AdamW optimizer; learning rate \(10^{-3}\); batch size 64; MSE loss; maximum 200 epochs; learning-rate reduction factor 0.5 with patience 10 and minimum learning rate \(10^{-5}\); early-stopping patience 30. \\

CSDI &
Imp./Fcst. &
Regularized values, masks, time embeddings, and laboratory embeddings &
Four residual layers; 64 residual channels; eight attention heads; diffusion-step, time, and laboratory embedding dimensions of 128, 128, and 16, respectively; 50 diffusion steps; quadratic noise schedule from \(\beta_1=10^{-4}\) to \(\beta_{50}=0.5\); Adam optimizer; learning rate \(10^{-3}\); batch sizes 16 and 8 and maximum epochs 200 and 100 for imputation and forecasting, respectively; 50 inference samples. \\

APN &
Fcst. &
Irregular context events and future queries &
Model dimension 16; time-encoding dimension 16; maximum 50 adaptive patches; dropout 0; Adam optimizer; learning rate 0.02; batch size 32; MSE loss; maximum 200 epochs; early-stopping patience 10; forecasting context and horizon of 84 and 42 steps, respectively. \\

TabPFN &
Imp./Fcst. &
Cell-wise row--column tabular examples &
Official pretrained \texttt{TabPFNRegressor}; ensemble size 10; optional additional mask and elapsed-time feature blocks disabled; no gradient-based parameter updates; independently conditioned on the visible cells of each admission. \\

\bottomrule
\end{tabular}
\caption{Baseline applicability, representations, and principal hyperparameters. Imp. and Fcst. denote imputation and forecasting, respectively. Unless otherwise stated, parameters not listed in the table retained their implementation defaults. All learned baselines were selected using only MIMIC-III training data and were applied to eICU without target-domain adaptation.}
\label{tab:baseline_hyperparameters}
\end{table*}

\subsection{Metrics}
\label{app:metrics}
We evaluate predictions only at ground-truth query locations and compute all metrics after inverse-transforming laboratory values to their original clinical scales. Let \(i\) index admissions and \(d\in\{1,\ldots,D\}\) index laboratory variables, with \(D=13\). For admission \(i\), define the query indices for laboratory \(d\) as
\begin{equation}
    \mathcal{Q}_{i,d}
    =
    \{\ell:(\ell,d)\in\mathcal{Q}_{i}\},
    \qquad
    N_d
    =
    \sum_i |\mathcal{Q}_{i,d}|,
    \label{eq:eval_query_set}
\end{equation}
where \(y_{i\ell d}\) and \(\widehat{y}_{i\ell d}\) denote the ground-truth and predicted values, respectively, at query location \((\ell,d)\), both expressed in the original laboratory scale.

\paragraph{Range-normalized root-mean-square deviation.}
Our primary metric is the range-normalized root-mean-square deviation (nRMSD) used in the DACMI benchmark \cite{luo_evaluating_2022}. For each admission and laboratory, normalization uses the range of the originally recorded measurements before constructing the evaluation queries. Let

\begin{equation}
\begin{aligned}
    \mathcal{O}_{i,d}
    &=
    \{\ell : M_{i,\ell d}=1\},\\
    R_{i,d}
    &=
    \max_{\ell\in\mathcal{O}_{i,d}} y_{i\ell d}
    -
    \min_{\ell\in\mathcal{O}_{i,d}} y_{i\ell d}.
\end{aligned}
\label{eq:evaluation_range}
\end{equation}

The cohort inclusion criteria ensure that the required laboratory measurements are available and that \(R_{i,d}>0\) for evaluated admission--laboratory pairs. Following DACMI, the nRMSD for laboratory \(d\) is
\begin{equation}
    \operatorname{nRMSD}_{d}
    =
    \sqrt{
    \frac{1}{N_d}
    \sum_i
    \sum_{\ell\in\mathcal{Q}_{i,d}}
    \left(
    \frac{
    \widehat{y}_{i\ell d}-y_{i\ell d}
    }{
    R_{i,d}
    }
    \right)^2
    }.
    \label{eq:nrmsd_lab}
\end{equation}
Range normalization makes errors from laboratory variables with different numerical scales comparable; lower nRMSD indicates better predictive accuracy.

We summarize nRMSD across laboratories in two complementary ways. \emph{Macro nRMSD} assigns equal weight to each laboratory by first computing Eq.~\eqref{eq:nrmsd_lab} independently and then averaging:
\begin{equation}
    \operatorname{nRMSD}_{\mathrm{macro}}
    =
    \frac{1}{D}
    \sum_{d=1}^{D}
    \operatorname{nRMSD}_{d}.
    \label{eq:nrmsd_macro}
\end{equation}
In contrast, \emph{Overall nRMSD} pools all evaluated query points across laboratories after applying the corresponding admission--laboratory range normalization:
\begin{equation}
    \operatorname{nRMSD}_{\mathrm{overall}}
    =
    \sqrt{
    \frac{
    \displaystyle
    \sum_{d=1}^{D}
    \sum_i
    \sum_{\ell\in\mathcal{Q}_{i,d}}
    \left(
    \frac{
    \widehat{y}_{i\ell d}-y_{i\ell d}
    }{
    R_{i,d}
    }
    \right)^2
    }{
    \displaystyle
    \sum_{d=1}^{D}N_d
    }
    }.
    \label{eq:nrmsd_overall}
\end{equation}
Thus, Macro nRMSD gives every laboratory equal influence regardless of the number of evaluated targets, whereas Overall nRMSD weights laboratories implicitly according to the number of query points they contribute.

\paragraph{Per-laboratory MAE and RMSE.}
To complement the normalized metric with errors expressed directly in the original clinical units, we report mean absolute error (MAE) and root-mean-square error (RMSE) separately for each laboratory:
\begin{equation}
    \operatorname{MAE}_{d}
    =
    \frac{1}{N_d}
    \sum_i
    \sum_{\ell\in\mathcal{Q}_{i,d}}
    \left|
    \widehat{y}_{i\ell d}-y_{i\ell d}
    \right|,
    \label{eq:mae_lab}
\end{equation}
and
\begin{equation}
    \operatorname{RMSE}_{d}
    =
    \sqrt{
    \frac{1}{N_d}
    \sum_i
    \sum_{\ell\in\mathcal{Q}_{i,d}}
    \left(
    \widehat{y}_{i\ell d}-y_{i\ell d}
    \right)^2
    }.
    \label{eq:rmse_lab}
\end{equation}
Because different laboratory tests have different units and numerical ranges, MAE and RMSE are not averaged across laboratories. Instead, they are used to characterize laboratory-specific prediction errors alongside \(\operatorname{nRMSD}_{d}\).

\section{Additional Results}
\label{app:additional_results}

\subsection{Per-Variable Results}
\label{app:per_variable_results}
Tables~\ref{tab:app_mimic_internal_imputation}--\ref{tab:app_eicu_forecasting}
provide the complete laboratory-level results for the three evaluation settings.
Each method is reported using three subrows for MAE, RMSE, and nRMSD.
MAE and RMSE are reported in the original units of each laboratory and are
therefore not aggregated across variables, whereas nRMSD is additionally
summarized using the Overall and Macro definitions from
Appendix~\ref{app:metrics}. For compactness, laboratory columns are
abbreviated as Cl (chloride), K (potassium), HCO$_3$ (bicarbonate), Na
(sodium), Hct (hematocrit), Hgb (hemoglobin), MCV (mean corpuscular volume),
Plt (platelets), WBC (white blood cells), RDW (red cell distribution width),
BUN (blood urea nitrogen), Cr (creatinine), and Glu (glucose).

\subsection{Extended PFN Predictor Ablations}
\label{app:extended_predictor_ablations}

The ablations in the main paper examine the two central design principles of MA-PFN: learning the predictor from prior-generated tasks and preserving the joint dependence between laboratory values and observation patterns. We additionally study the internal design of the PFN predictor. In these experiments, the diffusion prior, synthetic trajectory pool, context--query construction, training budget, and all unmodified predictor settings are held fixed. Only the indicated predictor component is changed. All variants are evaluated on external eICU imputation and reported as Overall/Macro nRMSD. The results are reported in Table~\ref{tab:app_predictor_component_ablations}. The reference configuration is the selected model in Table~\ref{tab:ours_hyperparameters}, which obtains (0.204/0.198).

\begin{table*}[t]
\centering
\begin{tabular}{@{}lcccc@{}}
\toprule
\textbf{Predictor variant}
& \textbf{Overall} $\downarrow$
& \textbf{Macro} $\downarrow$
& $\mathbf{\Delta}$ \textbf{Overall}
& $\mathbf{\Delta}$ \textbf{Macro} \\
\midrule

\multicolumn{5}{l}{\textit{Temporal inputs}} \\
Absolute time + delta time (default)
& 0.204 & 0.198 & --- & --- \\
w/o delta time
& 0.210 & 0.203 & +0.006 & +0.005 \\
w/o absolute time
& 0.205 & 0.199 & +0.001 & +0.001 \\
w/o absolute and delta time
& 0.219 & 0.213 & +0.015 & +0.015 \\

\midrule
\multicolumn{5}{l}{\textit{Missingness-aware auxiliary features}} \\
Mask convolution + laboratory statistics (default)
& 0.204 & 0.198 & --- & --- \\
w/o mask convolution
& 0.206 & 0.199 & +0.002 & +0.001 \\
w/o mask convolution and laboratory statistics
& 0.207 & 0.200 & +0.003 & +0.002 \\

\midrule
\multicolumn{5}{l}{\textit{Auxiliary-feature fusion operator}} \\
Concatenation (default)
& 0.204 & 0.198 & --- & --- \\
Addition
& 0.207 & 0.200 & +0.003 & +0.002 \\

\midrule
\multicolumn{5}{l}{\textit{Auxiliary-feature fusion stage}} \\
Before Transformer and prediction heads (default)
& 0.204 & 0.198 & --- & --- \\
Before Transformer only
& 0.209 & 0.203 & +0.005 & +0.005 \\
Before prediction heads only
& 0.207 & 0.201 & +0.003 & +0.003 \\

\bottomrule
\end{tabular}
\caption{Extended ablations of the PFN predictor on external eICU imputation. All results are Overall/Macro nRMSD. Each group changes only the indicated component relative to the selected configuration in Table~\ref{tab:ours_hyperparameters}. The $\Delta$) columns report absolute changes relative to the corresponding default, with positive values indicating degradation.}
\label{tab:app_predictor_component_ablations}
\end{table*}

\paragraph{Temporal representation.}
The temporal-feature ablations show that elapsed time since the most recent observation is the more important of the two temporal signals. Removing delta time increases Overall/Macro nRMSD from (0.204/0.198) to (0.210/0.203), whereas removing absolute time alone produces only a (0.001) increase in both metrics. This difference is consistent with the role of delta time in irregular clinical data: it directly encodes measurement recency and monitoring intensity, both of which can convey information about patient state and clinical concern.

Nevertheless, absolute time is not entirely redundant. Removing both temporal inputs produces a considerably larger degradation to (0.219/0.213), corresponding to relative increases of approximately (7.4\%) and (7.6\%) over the full predictor. The joint decline is substantially larger than that caused by removing either input individually. This suggests that the two temporal representations provide complementary reference frames: absolute time locates an observation within the admission, while delta time describes the local measurement process for a particular laboratory. The predictor can partially compensate when one reference frame remains available, but not when temporal information is removed altogether.

\paragraph{Missingness-aware auxiliary representations.}
Removing the local mask convolution increases Overall/Macro nRMSD to (0.206/0.199). Removing both the mask convolution and the per-laboratory statistics further increases error to (0.207/0.200). Although these changes are smaller than those caused by removing temporal information, they are consistent across Overall and Macro aggregation. The local convolution and summary statistics provide information at different scales: the convolution describes short-range measurement patterns surrounding each location, whereas the laboratory statistics summarize the amount and distribution of visible evidence over the admission. Their cumulative effect indicates that explicitly structured missingness features remain useful even though the raw observation mask is already included in the base token representation.

\paragraph{Fusion operator.}
Concatenating auxiliary features performs better than adding them to the base token representation. Replacing concatenation with addition increases Overall/Macro nRMSD from (0.204/0.198) to (0.207/0.200). Concatenation preserves separate subspaces for laboratory values, temporal information, laboratory-level statistics, and local mask patterns, allowing the fusion MLP to learn how strongly each source should contribute. Addition instead requires heterogeneous feature types to share the same representational coordinates before their interactions have been learned, potentially discarding information through premature mixing.

\paragraph{Fusion stage.}
The auxiliary representations are most effective when supplied both before contextual encoding and again before the prediction heads. Retaining only pre-Transformer fusion increases Overall/Macro nRMSD to (0.209/0.203), while retaining only pre-head fusion gives (0.207/0.201). The better result of pre-head-only fusion suggests that direct access to missingness-aware features at prediction time is particularly valuable and that auxiliary information introduced only at the input may be attenuated as it passes through multiple Transformer blocks. However, late fusion alone also underperforms the full configuration. The two-stage design therefore serves complementary roles: early fusion allows temporal and missingness information to influence contextual interactions across laboratories and time, while late fusion preserves direct access to these signals when producing query-specific predictions.

Taken together, these experiments indicate that the predictor's improvement does not arise from a single auxiliary feature. The largest component-level effect comes from explicitly representing irregular timing, particularly measurement recency. Local missingness features, laboratory statistics, and their placement at multiple stages provide smaller but cumulative improvements. The results also favor preserving heterogeneous information through concatenation rather than forcing it into a shared additive representation.

\subsection{Hyperparameter Sensitivity}
\label{app:pfn_hyperparameter_sensitivity}

We next examine whether external performance depends strongly on the principal architecture and optimization hyperparameters of the PFN predictor. All remaining settings are fixed to the selected configuration in Table~\ref{tab:ours_hyperparameters}. The tested settings and external eICU imputation results are shown in Table~\ref{tab:app_pfn_hyperparameter_sensitivity}; ($\dagger$) denotes the configuration used for the main experiments.

\begin{table*}[t]
\centering
\begin{tabular}{@{}lccc@{}}
\toprule
\textbf{Hyperparameter setting}
& \textbf{Value}
& \textbf{Overall} $\downarrow$
& \textbf{Macro} $\downarrow$ \\
\midrule

\multicolumn{4}{l}{\textit{Model dimension (d)}} \\
& 64
& 0.213 & 0.207 \\
& 128
& 0.206 & 0.199 \\
& $256^{\dagger}$
& 0.204 & 0.198 \\

\midrule
\multicolumn{4}{l}{\textit{\begin{tabular}[t]{@{}l@{}}
Number of 2D Transformer\\
blocks
\end{tabular}}} \\
& $(4^{\dagger})$
& 0.204 & 0.198 \\
& 8
& 0.205 & 0.199 \\
& 12
& 0.203 & 0.198 \\

\midrule
\multicolumn{4}{l}{\textit{Batch size}} \\
& 128
& 0.204 & 0.198 \\
& $(256^{\dagger})$
& 0.204 & 0.198 \\
& 512
& 0.205 & 0.198 \\

\midrule
\multicolumn{4}{l}{\textit{Initial learning rate}} \\
& $(1\times10^{-4})$
& 0.204 & 0.198 \\
& $(5\times10^{-4\dagger})$
& 0.204 & 0.198 \\
& $(1\times10^{-3})$
& 0.204 & 0.198 \\

\bottomrule
\end{tabular}
\caption{Sensitivity of the PFN predictor to model capacity and optimization hyperparameters on external eICU imputation. Results are reported as Overall/Macro nRMSD. The symbol ($\dagger$) marks the selected configuration used in the main experiments. Each experiment changes only the listed hyperparameter while retaining the remaining settings from Table~\ref{tab:ours_hyperparameters}.}
\label{tab:app_pfn_hyperparameter_sensitivity}
\end{table*}

\paragraph{Model width.}
Performance improves consistently as the model dimension increases. Increasing (d) from 64 to 128 reduces Overall/Macro nRMSD from (0.213/0.207) to (0.206/0.199), while increasing it further to 256 yields (0.204/0.198). Thus, insufficient token capacity has a clear adverse effect, particularly for (d=64), but the gain diminishes substantially between 128 and 256. This pattern suggests that the predictor requires enough width to jointly represent laboratory identity, values, temporal context, and missingness structure, after which additional capacity begins to saturate.

\paragraph{Transformer depth.}
Increasing the number of factorized 2D Transformer blocks does not yield a monotonic improvement. Four, eight, and twelve blocks obtain (0.204/0.198), (0.205/0.199), and (0.203/0.198), respectively. The twelve-block model improves Overall nRMSD by only (0.001) and leaves Macro nRMSD unchanged relative to four blocks, while eight blocks are marginally worse. The selected four-block architecture therefore provides a preferable accuracy--efficiency trade-off rather than relying on unnecessary depth.

\paragraph{Optimization settings.}
The predictor is insensitive to the tested batch sizes and learning rates. Batch sizes of 128 and 256 produce identical (0.204/0.198) results, while increasing the batch size to 512 changes only Overall nRMSD by (0.001). Similarly, learning rates from $(1 \times 10^{-4})$ to $(1 \times 10^{-3})$ produce identical results at three-decimal precision.

Overall, the sensitivity analysis reveals a specific capacity pattern. Adequate representational width is important, but performance is largely stable once the model reaches moderate width; increasing Transformer depth provides negligible benefit. In contrast, batch size and learning rate have almost no effect within the tested ranges. Combined with Table~\ref{tab:app_predictor_component_ablations}, these results suggest that the predictor benefits more from retaining clinically meaningful temporal and missingness information than from increasing architectural depth or narrowly tuning the optimizer.


\subsection{Quality of Synthetic Samples}
\label{app:synthetic_sample_quality}

We assess the fidelity of the learned diffusion prior from two complementary perspectives. First, we compare the marginal distributions of individual laboratory values. Second, we use Uniform Manifold Approximation and Projection (UMAP) to qualitatively examine whether real and synthetic trajectories occupy similar regions of the joint value--missingness representation space.

\begin{figure*}[tb]
\centering
\includegraphics[width=\textwidth]{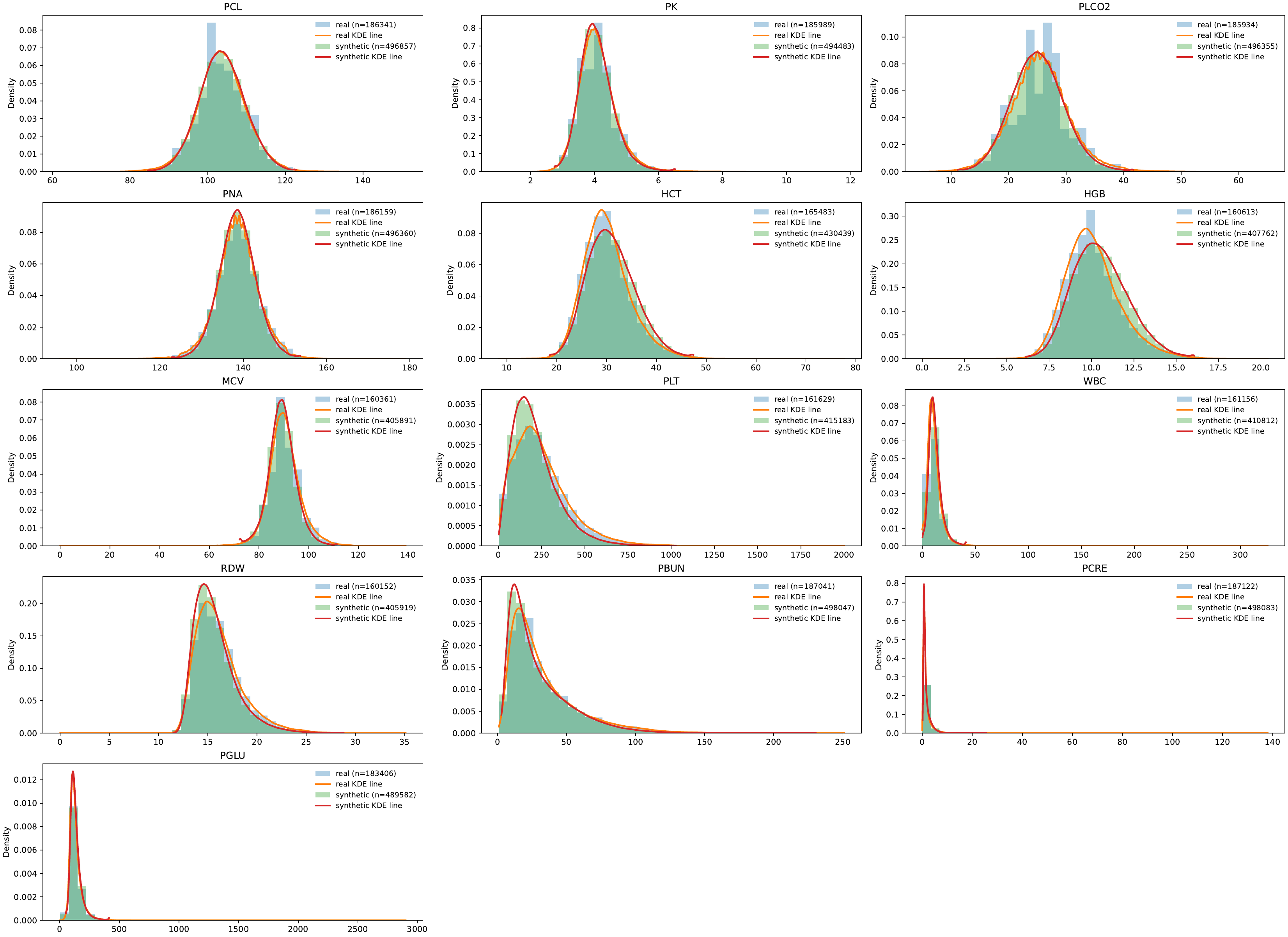}
\caption{Comparison of real and synthetic marginal laboratory-value distributions. For each of the 13 laboratory variables, all observed values are pooled across admissions and time steps after excluding unobserved entries using the corresponding masks. Normalized histograms and kernel density estimate (KDE) curves are shown using shared bin edges for the real and synthetic samples. The generated samples reproduce the central tendency and overall distributional shape of most laboratory variables, with the largest differences occurring in highly skewed or heavy-tailed variables.}
\label{fig:lab_distributions}
\end{figure*}

\paragraph{Marginal laboratory-value distributions.}
For each of the 13 laboratory variables, we extract all \emph{observed} values by excluding unobserved positions according to the corresponding masks and pooling the remaining values across admissions and time steps. We apply this procedure separately to the real training trajectories and the generated synthetic trajectories. Figure~\ref{fig:lab_distributions} presents the resulting normalized histograms together with kernel density estimate (KDE) curves.

Within each laboratory, the real and synthetic distributions are plotted using the same bin edges, making their locations, spreads, and shapes directly comparable. Overall, the generated samples closely follow the real distributions for most laboratory variables. In particular, the synthetic data generally preserve the position of the main mode, the dispersion around the center, and the overall distributional shape. This agreement is especially visible for relatively concentrated variables such as PCL, PK, PNA, MCV, and RDW, for which the real and synthetic KDE curves substantially overlap. For HCT, HGB, and PLCO2, the synthetic distributions also remain well aligned with the real data, with only minor differences in peak height or tail thickness.

More visible discrepancies occur for highly skewed or heavy-tailed variables. This is particularly noticeable for PLT, PBUN, and PCRE, whose empirical distributions contain pronounced right tails and comparatively rare extreme values. For these variables, the synthetic samples reproduce the main support and overall asymmetry of the real distributions, but differences remain in some low-density tail regions. These results indicate that the diffusion prior captures the dominant marginal characteristics of the laboratory values while reproducing rare extremes less precisely.

\paragraph{Joint trajectory structure.}
Marginal distributions alone do not determine whether the generator captures the multivariate temporal and missingness structure of complete admissions. We therefore complement the laboratory-wise analysis with a UMAP visualization. We randomly subsample 500 real and 500 synthetic trajectories and retain the first week of each trajectory, corresponding to the first 42 four-hour time steps. Each trajectory is represented using all 26 channels, comprising the 13 laboratory-value channels and their 13 corresponding mask channels. The value--mask representation is flattened across the temporal and variable dimensions, and the real and synthetic samples are jointly embedded into the same two-dimensional UMAP space.

\begin{figure}[tb]
\centering
\includegraphics[width=\columnwidth]{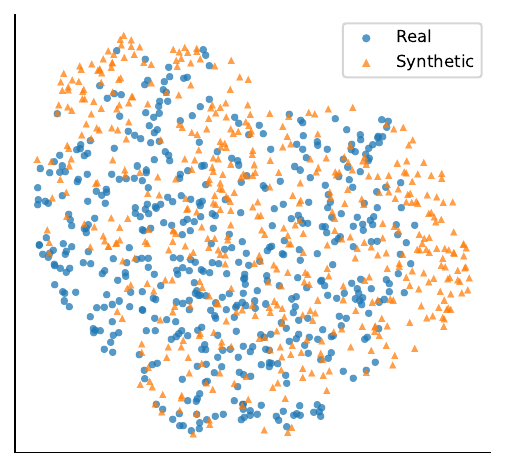}
\caption{UMAP visualization of real and synthetic clinical trajectories. The embedding uses the first week of data, corresponding to 42 four-hour time steps, from randomly selected subsets of 500 real and 500 synthetic trajectories. Each trajectory is represented using all 26 laboratory-value and mask channels, and both groups are jointly embedded in the same two-dimensional space. The substantial overlap and intermixing indicate that the generated trajectories reproduce much of the joint value--missingness structure of the real data, although modest differences in density remain in some peripheral regions.}
\label{fig:umap_real_synthetic}
\end{figure}

As shown in Figure~\ref{fig:umap_real_synthetic}, the real and synthetic trajectories occupy largely overlapping regions of the embedding. The two groups are extensively intermixed throughout the central area and across most of the surrounding structure, and there is no clearly separated cluster containing only real or only synthetic samples. This broad overlap suggests that the diffusion prior captures substantial aspects of the joint relationships among laboratory values, temporal evolution, and observation patterns during the first week.

Some localized differences nevertheless remain. Synthetic trajectories appear somewhat more frequently in parts of the upper and right periphery, whereas real trajectories are relatively more concentrated in portions of the lower region. These differences indicate that the global support is similar while the relative density assigned to particular trajectory patterns is not identical. Because UMAP is a nonlinear qualitative visualization whose geometry can depend on preprocessing and embedding hyperparameters, the figure should not be interpreted as evidence that the two distributions are equivalent. Rather, it provides complementary evidence that the synthetic trajectories reproduce the principal joint structure of the real data without forming a distinct synthetic-only manifold.

\subsection{Training and Inference Efficiency}
\label{app:efficiency}
We separately evaluate the one-time offline training cost and the prediction-time cost of the compared methods. The computational comparison is restricted to the imputation setting, in which all included methods are evaluated under the same query protocol. Forecasting runtimes are not reported separately because they follow the corresponding trained models' prediction pipelines and depend primarily on the number of future query locations.

\paragraph{Offline training time.}
Table~\ref{tab:imputation_training_time_detailed} reports the total wall-clock time required to obtain each model used in the final experiments. All measurements correspond to the final configurations used to produce the accuracy results in the main paper, rather than reduced configurations introduced solely for runtime evaluation. Each trainable baseline was executed 10 times using the same hyperparameters, data split, stopping criterion, and computational environment. We report the mean and standard deviation of the resulting wall-clock times.

Training time was measured from model initialization to creation of the final checkpoint and excludes dataset extraction, preprocessing, and disk transfer. When early stopping was used, the reported time includes the complete optimization procedure up to the selected stopping point. For XGBoost, the reported value includes fitting all 13 laboratory-specific regressors.

The MA-PFN offline pipeline is divided into three components. We separately report (i) training the missingness-aware diffusion prior, (ii) generating the fixed collection of 256{,}000 synthetic trajectories used for prior-data fitting, and (iii) training the PFN predictor on the resulting synthetic tasks. The total MA-PFN offline time is computed for each repetition by summing these three measured components.

Mean and linear interpolation require no parameter learning beyond the common preprocessing statistics and therefore have no offline training cost. TabPFN uses an externally pretrained checkpoint; consequently, its original pretraining cost was neither incurred nor measured in our experiments. Its admission-specific contextual prediction is included in the inference-time analysis.

\begin{table*}[tb]
\centering
\small
\setlength{\tabcolsep}{6pt}
\renewcommand{\arraystretch}{1.12}
\begin{tabular}{@{}p{2.0cm}p{9.2cm}p{3.0cm}@{}}
\toprule
\textbf{Method} &
\textbf{Offline component} &
\textbf{Training time (h)} \\
\midrule

Mean &
None &
-- \\

LI &
None &
-- \\

XGBoost &
Training 13 laboratory-specific regressors &
$\texttt{0.31} \pm \texttt{0.05}$ \\

GRU-D &
End-to-end training of the bidirectional recurrent model &
$\texttt{6.41} \pm \texttt{0.23}$ \\

GraFITi &
End-to-end training of the graph-based irregular time-series model &
$\texttt{0.73} \pm \texttt{0.07}$ \\

CSDI &
Training the conditional score-based diffusion model &
$\texttt{2.95} \pm \texttt{0.18}$ \\

TabPFN &
Use of an externally pretrained checkpoint; pretraining was not performed in this study &
N/A \\

\midrule

MA-PFN &
Training the missingness-aware diffusion prior &
$\texttt{14.19} \pm \texttt{0.94}$ \\

MA-PFN &
Generating 256{,}000 synthetic trajectories using the trained diffusion prior &
$\texttt{21.30} \pm \texttt{1.45}$ \\

MA-PFN &
Prior-data fitting of the PFN predictor using synthetic context--query tasks &
$\texttt{17.82} \pm \texttt{1.74}$ \\

\addlinespace[2pt]
\textbf{MA-PFN} &
\textbf{Total offline pipeline} &
$\mathbf{\texttt{53.31} \pm \texttt{1.26}}$ \\

\bottomrule
\end{tabular}
\caption{Offline training time for the imputation methods. Values denote wall-clock time in hours and are reported as mean $\pm$ standard deviation over 10 independent runs of the final configurations used to obtain the reported prediction results. For MA-PFN, training the diffusion prior, generating 256{,}000 synthetic trajectories, and prior-data fitting of the PFN predictor are reported separately. The total MA-PFN time is obtained by summing the three components within each run and subsequently computing the mean and standard deviation across the resulting run-level totals. Dataset extraction, preprocessing, and disk input/output times are excluded. Mean and linear interpolation (LI) require no offline training. TabPFN uses an externally pretrained checkpoint; therefore, its original pretraining cost was not measured in this study.}
\label{tab:imputation_training_time_detailed}
\end{table*}

\begin{figure}[tb]
\centering
\includegraphics[width=\columnwidth]{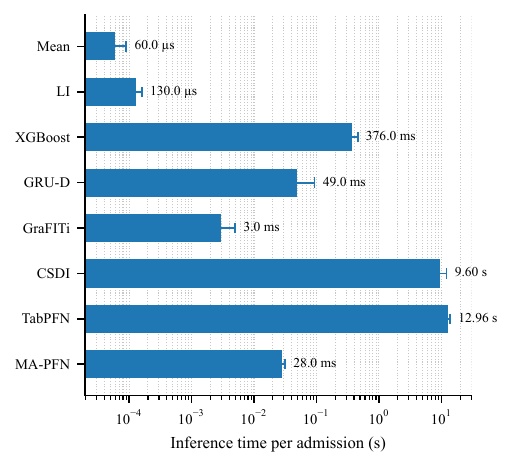}
\caption{Inference latency for imputation on the eICU cohort. Bars show the mean end-to-end inference time per admission, and error bars show one standard deviation across admissions. The horizontal axis uses a logarithmic scale. Measurements use batch size one after model warm-up and exclude model loading, dataset loading, and disk input/output, while including context--query construction, required sliding-window processing, and prediction aggregation. All timings use the final configurations employed for the reported accuracy results. CSDI latency includes 50 conditional samples, whereas MA-PFN inference consists only of a PFN forward pass and does not execute the diffusion prior.}
\label{fig:inference_time}
\end{figure}

\paragraph{Inference time.}
Figure~\ref{fig:inference_time} compares end-to-end imputation latency on the eICU cohort. For each method, the bar reports the mean inference time per admission, and the error bar represents one standard deviation across admissions. The standard deviation reflects variation in computational cost caused by differences in admission length, observation density, and the number of model-specific operations required for individual admissions.

All measurements correspond to the final model configurations used in the accuracy experiments. Models were loaded before timing, and warm-up predictions were performed before collecting measurements. The timed region excludes model loading, dataset loading, and disk input/output, but includes construction of model-specific inputs, context--query preparation, padding, required sliding-window evaluation, and aggregation of overlapping predictions. GPU operations were explicitly synchronized before and after each timed prediction. Latency was measured using a batch size of one to represent prediction for an individual admission.

For CSDI, inference time includes generation of the same 50 conditional samples used to compute the reported prediction results. For TabPFN, it includes constructing the admission-specific in-context regression dataset and running the pretrained model, but no gradient-based parameter updates. For MA-PFN, inference uses only the trained PFN predictor; the missingness-aware diffusion model is used exclusively during offline prior learning and is not executed for target admissions.

The offline and online measurements emphasize the amortized nature of MA-PFN. Although learning the diffusion prior and fitting the PFN predictor incur a one-time offline cost, the learned prior is compiled into the predictor parameters. Prediction for a new admission therefore requires neither diffusion sampling nor target-domain parameter optimization.



\section{Limitations and Future Directions}
\label{app:limitations_future}

\subsection{Limitations}
\paragraph{Restricted clinical variable set.}
The present study considers 13 numerical laboratory variables shared between MIMIC-III and eICU. This controlled setting allows us to study irregular sampling and informative missingness consistently across the two databases, but represents only a small subset of the information available in an EHR. In particular, the current model does not incorporate vital signs, medications, diagnoses, procedures, static patient characteristics, or categorical clinical events. Consequently, the results establish the feasibility of missingness-aware prior-data fitting for laboratory trajectories rather than general-purpose modeling of complete patient records.

\paragraph{Limited external validation.}
The learned prior and predictor are developed exclusively from MIMIC-III, with eICU reserved as an independent external domain. This design provides a direct test of transfer without target-domain training, preprocessing-statistic estimation, hyperparameter selection, or parameter adaptation. Nevertheless, a single external database cannot characterize all clinically relevant forms of distribution shift. Hospitals may differ in patient populations, laboratory assays, care pathways, test-ordering policies, documentation practices, and temporal periods. Validation across additional institutions, healthcare systems, countries, and temporally separated cohorts is therefore required before drawing broader conclusions about cross-domain robustness.

\paragraph{Dependence on the source-domain prior.}
Prior-data fitting does not make MA-PFN independent of its source data. The diffusion model learns its task prior from the MIMIC-III training distribution, and the PFN subsequently learns an inference procedure from tasks generated by that prior. Physiological states, laboratory relationships, monitoring policies, or rare clinical trajectories that are absent or poorly represented in MIMIC-III may therefore also be absent or underrepresented in the synthetic task distribution. The behavior of MA-PFN under domain shift consequently depends on how well the learned prior covers the mechanisms encountered in the target population.

\paragraph{Dependence on synthetic-prior fidelity.}
The downstream predictor is trained on trajectories generated by the diffusion model rather than directly on real admissions. Imperfect reproduction of laboratory distributions, temporal dependencies, observation frequencies, rare trajectories, or value--mask coupling can therefore propagate into the context--query tasks used for prior-data fitting. Generating a large number of synthetic trajectories increases the diversity of sampled tasks but cannot recover clinical mechanisms that the generative prior itself has failed to learn. The quality and coverage of the learned prior consequently constitute an important upper bound on the usefulness of subsequent prior-data fitting.

\paragraph{Temporal discretization.}
Clinical measurements are aligned to a 4-hour grid before modeling. The observation mask and recurrence features retain information about whether and how recently each laboratory was observed, but the representation no longer preserves exact continuous chart times within each bin. When a laboratory is measured multiple times within the same interval, only the latest measurement is retained. MA-PFN therefore models irregular clinical trajectories at the selected temporal resolution rather than directly representing the original asynchronous event sequence. Different resolutions, adaptive discretization, or continuous-time representations may preserve finer temporal structure.

\paragraph{Evaluation of imputation targets.}
Naturally missing laboratory values have no observed ground truth and therefore cannot be evaluated directly. Both internal and external imputation evaluation instead hide measurements that were originally observed and ask the model to reconstruct them from the remaining context. This protocol permits objective comparison across methods and follows the task construction described in Sec.~\ref{app:task_construction}, but performance on artificially withheld observations does not prove equivalent accuracy for values that were naturally unobserved. This distinction is particularly relevant under informative or MNAR-like missingness, because naturally missing locations may systematically differ from locations at which clinicians chose to obtain measurements.

\paragraph{Forecasting cohort selection.}
The forecasting experiment requires 14 days of available context and evaluates predictions over the subsequent week. Admissions shorter than two weeks are therefore excluded, and the eICU forecasting cohort additionally requires at least 13 observed laboratory measurements in the forecast horizon. These criteria ensure that forecasting examples contain sufficient history and evaluable future targets, but they preferentially select longer and sufficiently monitored admissions. Forecasting results should consequently not be interpreted as representative of short ICU stays or patients with extremely sparse future laboratory monitoring.

\paragraph{Retrospective evaluation and clinical utility.}
The experiments evaluate reconstruction and forecasting accuracy retrospectively on existing EHR databases. They do not establish that the predictions improve clinical decision-making, workflow, treatment selection, or patient outcomes. Clinical deployment would additionally require prospective evaluation, analysis of clinically consequential errors, assessment across relevant patient subgroups and care settings, and investigation of how predictions interact with clinician decisions. MA-PFN should therefore be interpreted as a predictive modeling framework rather than a validated clinical decision-support system.

\paragraph{No formal privacy guarantee.}
Training the PFN on synthetic trajectories does not by itself provide a formal privacy guarantee. The diffusion prior is learned from real source-domain admissions, and generative models can in principle reproduce or memorize properties of their training data. Although synthetic prior-data fitting changes how the downstream predictor is trained, privacy must be evaluated independently. Membership-inference analyses, similarity-based audits, or training the generative prior under formal privacy mechanisms such as differential privacy would be required before synthetic trajectories could be interpreted as privacy-preserving data products.

\subsection{Future Directions}
\paragraph{Broader heterogeneous EHR modeling.}
A natural extension is to learn the prior over a substantially richer clinical state. In addition to the laboratory variables considered here, future models could incorporate vital signs, medications, diagnoses, procedures, interventions, demographics, and other static or longitudinal information. The mixed continuous--discrete formulation provides a natural starting point: numerical measurements can be modeled through continuous diffusion, while binary or categorical clinical variables can be represented through discrete diffusion. This would allow the task prior to capture dependencies across multiple clinical modalities rather than laboratory physiology alone.

\paragraph{Multi-institutional task priors.}
The current prior reflects one source database. Future work could instead learn a prior from multiple healthcare systems so that the synthetic task distribution spans a broader range of populations, clinical practices, measurement frequencies, and observation mechanisms. Such a design could reduce dependence on any single institution and provide a stronger basis for zero-adaptation transfer. In this setting, databases currently used for external evaluation, such as eICU, should only be incorporated into prior learning under a redesigned evaluation protocol that reserves separate unseen institutions or datasets for external validation.

\paragraph{Finer and continuous temporal representations.}
The current 4-hour grid provides a practical common coordinate system for irregular laboratory trajectories. Future work could investigate adaptive temporal bins, event-based representations, or continuous-time components that retain exact measurement times while preserving explicit observation-process modeling. Such representations may be particularly useful when extending MA-PFN to variables whose clinically relevant dynamics occur at substantially different temporal scales.

\paragraph{Broader context--query task families.}
The current prior-data fitting procedure focuses on laboratory imputation and forecasting. The same principle can expose a predictor to a broader distribution of clinical inference tasks by varying the context, query variables, and prediction horizons during prior fitting. Potential extensions include multi-horizon forecasting, cross-variable prediction, conditional trajectory generation, phenotype prediction, and risk estimation. Learning across a heterogeneous family of tasks would further test whether a common clinical prior can support reusable inference rather than separate models specialized to individual prediction problems.

\paragraph{Toward an EHR foundation model.}
The present model should be viewed as a step toward, rather than an instance of, a general-purpose EHR foundation model. Scaling the learned prior to heterogeneous clinical variables, larger multi-institutional datasets, diverse observation mechanisms, and a broader family of context--query tasks could produce a reusable prior-fitted model whose inference procedure transfers across clinical prediction problems without conventional task-specific retraining. Establishing this direction will require substantially broader pretraining data and evaluation across unseen institutions, patient populations, variable sets, and downstream tasks.

\begin{table*}[t]
\centering
\scriptsize
\setlength{\tabcolsep}{1.0mm}
\begin{tabular}{@{}ll*{15}{c}@{}}
\toprule
& &
\multicolumn{13}{c}{\textbf{Laboratory-Specific Error}} &
\multicolumn{2}{c}{\textbf{Aggregate nRMSD}} \\
\cmidrule(lr){3-15}
\cmidrule(lr){16-17}

\textbf{Method} & \textbf{Metric}
& \textbf{Cl} & \textbf{K} & \textbf{HCO$_3$} & \textbf{Na}
& \textbf{Hct} & \textbf{Hgb} & \textbf{MCV} & \textbf{Plt}
& \textbf{WBC} & \textbf{RDW} & \textbf{BUN} & \textbf{Cr}
& \textbf{Glu} & \textbf{Overall} & \textbf{Macro} \\

\midrule
\multicolumn{2}{r}{\textbf{No. Queries}}
& 8,267 & 8,267 & 8,267 & 8,267 & 8,267 & 8,267 & 8,267 & 8,267 & 8,267 & 8,267 & 8,267 & 8,267 & 8,267
& 107,471 & \textemdash \\
\midrule

Mean
& MAE   & 3.199 & 0.390 & 2.657 & 2.610 & 2.645 & 0.881 & 1.637 & 66.69 & 2.946 & 0.570 & 8.427 & 0.330 & 31.14 & \textemdash & \textemdash \\
& RMSE  & 4.225 & 0.543 & 3.453 & 3.488 & 3.544 & 1.173 & 2.292 & 100.2 & \underline{4.843} & 0.876 & 13.32 & 0.641 & 54.30 & \textemdash & \textemdash \\
& nRMSD & 0.293 & 0.280 & 0.298 & 0.290 & 0.295 & 0.295 & 0.311 & 0.319 & 0.301 & 0.315 & 0.316 & 0.309 & 0.281 & 0.300 & 0.300 \\
\midrule

LI
& MAE   & 1.912 & 0.361 & 1.658 & 1.782 & 1.917 & 0.626 & 1.186 & \underline{25.62} & \underline{1.770} & 0.269 & 2.667 & 0.158 & 31.67 & \textemdash & \textemdash \\
& RMSE  & 2.741 & 0.546 & 2.258 & 2.447 & 2.757 & 0.907 & 1.760 & \underline{41.40} & \textbf{2.969} & \textbf{0.413} & 5.347 & 0.342 & 64.31 & \textemdash & \textemdash \\
& nRMSD & 0.223 & 0.281 & 0.231 & 0.246 & 0.251 & 0.249 & 0.284 & \underline{0.198} & \underline{0.240} & 0.226 & 0.175 & 0.233 & 0.299 & 0.243 & 0.241 \\
\midrule

XGBoost
& MAE   & 3.277 & 0.404 & 2.638 & 2.766 & 2.552 & 0.841 & 1.994 & 62.16 & 3.226 & 0.658 & 8.769 & 0.399 & 32.33 & \textemdash & \textemdash \\
& RMSE  & 4.341 & 0.560 & 3.450 & 3.682 & 3.345 & 1.106 & 2.784 & 96.45 & 7.206 & 1.034 & 14.45 & 0.836 & 58.66 & \textemdash & \textemdash \\
& nRMSD & 0.322 & 0.305 & 0.324 & 0.323 & 0.307 & 0.310 & 0.432 & 0.394 & 0.376 & 0.464 & 0.391 & 0.413 & 0.309 & 0.363 & 0.359 \\
\midrule

GRU-D
& MAE   & 1.434 & \underline{0.305} & 1.534 & 1.346 & 0.999 & 0.309 & 1.386 & 31.96 & 1.887 & 0.382 & 4.110 & 0.230 & \underline{26.93} & \textemdash & \textemdash \\
& RMSE  & 2.051 & 0.459 & \underline{2.076} & 1.845 & 1.527 & 0.461 & 2.045 & 60.43 & 5.512 & 0.629 & 7.482 & 0.597 & \underline{53.06} & \textemdash & \textemdash \\
& nRMSD & \underline{0.169} & \underline{0.242} & \underline{0.212} & 0.188 & 0.152 & 0.140 & 0.412 & 0.260 & 0.252 & 0.404 & 0.242 & 0.290 & \underline{0.265} & 0.261 & 0.248 \\
\midrule

GraFITi
& MAE   & \underline{1.423} & 0.317 & \underline{1.475} & \underline{1.299} & \underline{0.881} & \underline{0.234} & \textbf{0.942} & 29.39 & 2.012 & \textbf{0.254} & \underline{2.631} & \underline{0.129} & 28.55 & \textemdash & \textemdash \\
& RMSE  & \underline{2.038} & \underline{0.454} & 2.144 & \underline{1.774} & 1.487 & \underline{0.399} & \textbf{1.355} & 45.48 & 7.242 & 0.529 & \underline{4.218} & \underline{0.277} & 54.51 & \textemdash & \textemdash \\
& nRMSD & \underline{0.169} & 0.244 & \underline{0.211} & \underline{0.183} & \underline{0.141} & 0.\underline{118} & \textbf{0.262} & 0.215 & 0.249 & \textbf{0.210} & \underline{0.165} & \underline{0.218} & 0.280 & \underline{0.209} & \underline{0.205} \\
\midrule

CSDI
& MAE   & 1.796 & 0.389 & 1.931 & 1.685 & 0.973 & 0.294 & 1.527 & 34.27 & 2.431 & 0.401 & 3.700 & 0.203 & 32.22 & \textemdash & \textemdash \\
& RMSE  & 2.476 & 0.547 & 2.532 & 2.415 & \underline{1.466} & 0.437 & 2.627 & 50.78 & 17.91 & 0.671 & 5.955 & 0.423 & 58.46 & \textemdash & \textemdash \\
& nRMSD & 0.218 & 0.320 & 0.275 & 0.340 & 0.147 & 0.137 & 0.395 & 0.272 & 0.325 & 0.384 & 0.222 & 0.309 & 0.332 & 0.293 & 0.282 \\
\midrule

TabPFN
& MAE   & 3.429 & 0.440 & 2.98 & 2.839 & 2.69 & 0.879 & 2.162 & 66.24 & 3.412 & 0.691 & 8.991 & 0.423 & 34.63 & \textemdash & \textemdash \\
& RMSE  & 4.601 & 0.645 & 4.017 & 3.775 & 3.52 & 1.150 & 3.015 & 101.3 & 7.691 & 1.078 & 14.88 & 0.887 & 62.09 & \textemdash & \textemdash \\
& nRMSD & 0.337 & 0.355 & 0.354 & 0.349 & 0.319 & 0.322 & 0.467 & 0.414 & 0.397 & 0.484 &  0.399 & 0.438 & 0.331 & 0.391 & 0.382 \\
\midrule

\textbf{MA-PFN}
& MAE   & \textbf{1.254} & \textbf{0.296} & \textbf{1.366} & \textbf{1.189} & \textbf{0.631} & \textbf{0.196} & \underline{0.967} & \textbf{22.81} & \textbf{1.762} & \underline{0.260} & \textbf{2.407} & \textbf{0.119} & \textbf{25.81} & \textemdash & \textemdash \\
& RMSE  & \textbf{1.846} & \textbf{0.436} & \textbf{1.850} & \textbf{1.657} & \textbf{1.128} & \textbf{0.334} & \underline{1.370} & \textbf{40.58} & 6.255 & \underline{0.519} & \textbf{3.850} & \textbf{0.246} & \textbf{49.47} & \textemdash & \textemdash \\
& nRMSD & \textbf{0.150} & \textbf{0.235} & \textbf{0.193} & \textbf{0.169} & \textbf{0.107} & \textbf{0.099} & \underline{0.265} & \textbf{0.166} & \textbf{0.218} & \underline{0.214} & \textbf{0.151} & \textbf{0.201} & \textbf{0.254} & \textbf{0.193} & \textbf{0.186} \\
\bottomrule
\end{tabular}
\caption{Per-laboratory results for internal imputation on the MIMIC-III test set. ``No. Queries'' denotes the number of evaluated observed query points for each laboratory, with Overall reporting the total number of queries across all 13 laboratories. MAE and RMSE are reported in the original laboratory units; nRMSD is range-normalized. Overall nRMSD pools all evaluated queries, whereas Macro nRMSD gives equal weight to each laboratory. Lower values are better. Best results are shown in \textbf{bold} and second-best results are \underline{underlined}.}
\label{tab:app_mimic_internal_imputation}
\end{table*}

\begin{table*}[t]
\centering
\scriptsize
\setlength{\tabcolsep}{1.0mm}
\begin{tabular}{@{}ll*{15}{c}@{}}
\toprule
& &
\multicolumn{13}{c}{\textbf{Laboratory-Specific Error}} &
\multicolumn{2}{c}{\textbf{Aggregate nRMSD}} \\
\cmidrule(lr){3-15}
\cmidrule(lr){16-17}

\textbf{Method} & \textbf{Metric}
& \textbf{Cl} & \textbf{K} & \textbf{HCO$_3$} & \textbf{Na}
& \textbf{Hct} & \textbf{Hgb} & \textbf{MCV} & \textbf{Plt}
& \textbf{WBC} & \textbf{RDW} & \textbf{BUN} & \textbf{Cr}
& \textbf{Glu} & \textbf{Overall} & \textbf{Macro} \\

\midrule
\multicolumn{2}{r}{\textbf{No. Queries}}
& 21,423 & 21,423 & 21,423 & 21,423 & 21,423 & 21,423 & 21,423 & 21,423 & 21,423 & 21,423 & 21,423 & 21,423 & 21,423
& 278,499 & \textemdash \\
\midrule

Mean
& MAE   & 3.360 & 0.377 & 2.660 & 2.817 & 2.756 & 0.930 & 1.474 & 56.41 & 3.025 & 0.608 & 8.133 & 0.356 & 29.21 & \textemdash & \textemdash \\
& RMSE  & 4.411 & 0.498 & 3.483 & 3.726 & 3.742 & 1.271 & 2.070 & 81.92 & 4.357 & 0.950 & 11.73 & 0.651 & 44.08 & \textemdash & \textemdash \\
& nRMSD & 0.308 & 0.287 & 0.307 & 0.300 & 0.294 & 0.296 & 0.316 & 0.324 & 0.305 & 0.321 & 0.322 & 0.310 & 0.287 & 0.306 & 0.306 \\
\midrule

LI
& MAE   & 1.897 & \underline{0.310} & 1.689 & 1.735 & 1.817 & 0.605 & \underline{0.889} & \underline{21.37} & \underline{1.848} & \textbf{0.263} & 3.382 & \underline{0.185} & 26.44 & \textemdash & \textemdash \\
& RMSE  & 2.689 & 0.430 & 2.296 & 2.345 & 2.644 & 0.902 & \underline{1.354} & \underline{33.59} & \underline{2.938} & 0.447 & 6.218 & \underline{0.433} & 42.87 & \textemdash & \textemdash \\
& nRMSD & 0.223 & \underline{0.269} & 0.242 & 0.236 & 0.239 & 0.239 & \textbf{0.254} & \underline{0.197} & \underline{0.241} & \textbf{0.218} & 0.190 & \underline{0.231} & 0.291 & 0.237 & 0.236 \\
\midrule

XGBoost
& MAE   & 3.470 & 0.391 & 2.665 & 2.932 & 2.721 & 0.900 & 1.977 & 55.28 & 3.244 & 0.765 & 8.891 & 0.450 & 31.34 & \textemdash & \textemdash \\
& RMSE  & 4.543 & 0.518 & 3.476 & 3.882 & 3.596 & 1.194 & 2.765 & 80.14 & 4.781 & 1.248 & 13.32 & 0.903 & 48.40 & \textemdash & \textemdash \\
& nRMSD & 0.354 & 0.317 & 0.344 & 0.340 & 0.339 & 0.338 & 0.534 & 0.461 & 0.377 & 0.546 & 0.408 & 0.496 & 0.331 & 0.406 & 0.398 \\
\midrule

GRU-D
& MAE   & \underline{1.566} & \textbf{0.288} & 1.640 & \underline{1.525} & \underline{1.188} & 0.372 & 1.257 & 27.11 & 1.872 & 0.481 & 4.064 & 0.264 & 25.87 & \textemdash & \textemdash \\
& RMSE  & \underline{2.150} & \underline{0.392} & \underline{2.183} & \underline{2.060} & 1.853 & 0.586 & 1.935 & 44.77 & 2.990 & 0.872 & 6.593 & 0.608 & 42.11 & \textemdash & \textemdash \\
& nRMSD & \underline{0.190} & \textbf{0.245} & \underline{0.237} & \underline{0.208} & 0.177 & 0.162 & 0.487 & 0.270 & 0.264 & 0.476 & 0.237 & 0.342 & \underline{0.279} & 0.292 & 0.274 \\
\midrule

GraFITi
& MAE   & 1.729 & 0.301 & \underline{1.531} & 1.830 & 1.287 & \underline{0.353} & 1.112 & 22.58 & 1.989 & 0.298 & \underline{3.140} & 0.226 & \underline{24.99} & \textemdash & \textemdash \\
& RMSE  & 2.892 & 0.496 & 2.253 & 2.474 & \underline{1.839} & \underline{0.525} & 1.591 & 36.92 & 3.335 & 0.470 & \underline{5.927} & 0.492 & \underline{40.77} & \textemdash & \textemdash \\
& nRMSD & 0.215 & 0.279 & 0.248 & 0.249 & \underline{0.168} & \underline{0.155} & 0.289 & 0.199 & 0.249 & 0.239 & \underline{0.182} & 0.257 & 0.297 & 0.237 & 0.233 \\
\midrule

CSDI
& MAE   & 2.024 & 0.388 & 2.183 & 1.917 & \underline{1.182} & 0.365 & 1.481 & 33.79 & 2.376 & 0.464 & 4.313 & 0.242 & 32.59 & \textemdash & \textemdash \\
& RMSE  & 2.724 & 0.518 & 2.865 & 2.520 & 1.894 & 0.622 & 2.021 & 48.57 & 3.567 & 0.707 & 6.739 & 0.497 & 50.35 & \textemdash & \textemdash \\
& nRMSD & 0.266 & 0.350 & 0.330 & 0.273 & 0.188 & 0.175 & 0.513 & 0.320 & 0.323 & 0.423 & 0.261 & 0.349 & 0.373 & 0.321 & 0.319 \\
\midrule

TabPFN
& MAE   & 3.499 & 0.383 & 2.625 & 3.102 & 2.485 & 0.792 & 2.188 & 57.49 & 3.344 & 0.793 & 8.851 & 0.485 & 31.15 & \textemdash & \textemdash \\
& RMSE  & 4.536 & 0.581 & 3.416 & 3.915 & 3.216 & 1.283 & 2.820 & 83.90 & 4.881 & 1.292 & 13.62 & 0.944 & 47.86 & \textemdash & \textemdash \\
& nRMSD &  0.352 & 0.319 & 0.344 & 0.348 & 0.322 & 0.335 & 0.564 & 0.471 & 0.377 & 0.552 & 0.408 & 0.502 & 0.328 & 0.401 & 0.401 \\
\midrule

\textbf{MA-PFN}
& MAE  & \textbf{1.361} & \textbf{0.288} & \textbf{1.487} & \textbf{1.362} & \textbf{0.648} & \textbf{0.221} & \textbf{0.853} & \textbf{19.85} & \textbf{1.715} & \underline{0.283} & \textbf{2.718} & \textbf{0.135} & \textbf{24.80} & \textemdash & \textemdash \\
& RMSE  & \textbf{1.882} & \textbf{0.389} & \textbf{1.991} & \textbf{1.851} & \textbf{1.340} & \textbf{0.456} & \textbf{1.280} & \textbf{30.10} & \textbf{2.662} & \textbf{0.442} & \textbf{4.288} & \textbf{0.270} & \textbf{39.77} & \textemdash & \textemdash \\
& nRMSD & \textbf{0.169} & \textbf{0.245} & \textbf{0.214} & \textbf{0.190} & \textbf{0.112} & \textbf{0.111} & \underline{0.285} & \textbf{0.179} & \textbf{0.218} & \underline{0.220} & \textbf{0.163} & \textbf{0.202} & \textbf{0.268} & \textbf{0.204} & \textbf{0.198} \\
\bottomrule
\end{tabular}
\caption{Per-laboratory results for external imputation on eICU. All models are evaluated without target-domain training or adaptation. ``No. Queries'' denotes the number of evaluated observed query points for each laboratory, with Overall reporting the total number of queries across all 13 laboratories. MAE and RMSE are reported in the original laboratory units; lower values are better for all metrics. Best results are shown in \textbf{bold} and second-best results are \underline{underlined}.}
\label{tab:app_eicu_external_imputation}
\end{table*}

\begin{table*}[t]
\centering
\scriptsize
\setlength{\tabcolsep}{1.0mm}
\begin{tabular}{@{}ll*{15}{c}@{}}
\toprule
& &
\multicolumn{13}{c}{\textbf{Laboratory-Specific Error}} &
\multicolumn{2}{c}{\textbf{Aggregate nRMSD}} \\
\cmidrule(lr){3-15}
\cmidrule(lr){16-17}

\textbf{Method} & \textbf{Metric}
& \textbf{Cl} & \textbf{K} & \textbf{HCO$_3$} & \textbf{Na}
& \textbf{Hct} & \textbf{Hgb} & \textbf{MCV} & \textbf{Plt}
& \textbf{WBC} & \textbf{RDW} & \textbf{BUN} & \textbf{Cr}
& \textbf{Glu} & \textbf{Overall} & \textbf{Macro} \\

\midrule
\multicolumn{2}{r}{\textbf{No. Queries}}
& 46,575 & 53,394 & 46,445 & 50,257 & 45,293 & 46,429 & 42,142 & 42,424 & 41,718 & 41,961 & 45,373 & 46,895 & 48,233
& 597,139 & \textemdash \\
\midrule

Mean
& MAE   & 4.271 & 0.432 & 3.154 & 3.754 & 3.286 & 1.121 & 2.274 & 101.1 & 3.928 & 1.157 & 11.95 & 0.521 & 31.35 & \textemdash & \textemdash \\
& RMSE  & 5.500 & 0.572 & 4.094 & 4.928 & 4.290 & 1.467 & 3.128 & 144.7 & 5.452 & 1.734 & 16.71 & 0.895 & \underline{44.74} & \textemdash & \textemdash \\
& nRMSD & 0.346 & 0.293 & 0.333 & 0.332 & 0.312 & 0.310 & 0.382 & 0.445 & 0.336 & 0.428 & 0.380 & 0.348 & 0.283 & 0.349 & 0.348 \\
\midrule

LOCF
& MAE   & 3.558 & 0.457 & 2.775 & 3.309 & 2.712 & 0.902 & 1.657 & \textbf{59.12} & 3.329 & \textbf{0.699} & 9.33 & 0.391 & 34.95 & \textemdash & \textemdash \\
& RMSE  & 4.904 & 0.612 & 3.742 & 4.546 & 3.722 & 1.243 & 2.462 & \textbf{88.19} & 4.963 & \textbf{1.120} & 14.55 & 0.769 & 54.63 & \textemdash & \textemdash \\
& nRMSD & 0.323 & 0.322 & 0.322 & 0.324 & 0.293 & 0.288 & 0.332 & \textbf{0.331} & 0.327 & \textbf{0.328} & 0.336 & 0.312 & 0.326 & 0.320 & 0.320 \\
\midrule

XGBoost
& MAE   & 3.575 & 0.405 & 2.801 & 3.202 & 2.681 & 0.906 & 2.458 & 72.88 & \underline{3.186} & 0.964 & 10.22 & 0.523 & 29.70 & \textemdash & \textemdash \\
& RMSE  & 4.683 & 0.531 & 3.649 & 4.264 & 3.513 & 1.180 & 3.457 & 102.5 & 4.738 & 1.490 & 15.21 & 1.037 & 45.72 & \textemdash & \textemdash \\
& nRMSD & 0.329 & 0.284 & 0.326 & 0.312 & 0.301 & 0.298 & 0.623 & 0.499 & 0.332 & 0.513 & 0.380 & 0.425 & 0.287 & 0.385 & 0.377 \\
\midrule

GRU-D 
& MAE   & \textbf{3.225} & 0.386 & \textbf{2.557} & 2.965 & \textbf{2.536} & \textbf{0.858} & 1.975 & 64.90 & \textbf{2.940} & 0.935 & 8.977 & 0.462 & 29.48 & \textemdash & \textemdash \\
& RMSE  & \textbf{4.235} & \underline{0.508} & \underline{3.355} & 3.975 & \textbf{3.303} & \textbf{1.102} & 2.670 & \underline{91.98} & \underline{4.456} & 1.427 & \underline{13.12} & 0.918 & 45.94 & \textemdash & \textemdash \\
& nRMSD & \underline{0.295} & 0.271 & 0.299 & 0.290 & 0.287 & 0.282 & 0.502 & 0.404 & \underline{0.309} & 0.506 & 0.340 & 0.375 & 0.281 & 0.346 & 0.341 \\
\midrule

GraFITi 
& MAE   & 3.439 & 0.400 & \underline{2.588} & 2.994 & 2.703 & 0.873 & \textbf{1.507} & 65.83 & 3.354 & 0.764 & 9.052 & \underline{0.382} & \underline{29.06} & \textemdash & \textemdash \\
& RMSE  & 5.315 & 0.567 & 4.016 & 4.634 & 3.872 & 1.294 & 2.876 & 93.62 & 5.706 & 1.288 & 14.86 & 0.862 & 47.87 & \textemdash & \textemdash \\
& nRMSD &  0.299 & 0.281 & \underline{0.291} & 0.288 & 0.296 & 0.287 & 0.343 & 0.363 & 0.311 & 0.378 & \textbf{0.305} & 0.310 & \underline{0.277} & 0.312 & 0.310 \\
\midrule

APN 
& MAE   & 3.280 & \textbf{0.376} & 2.781 &  \textbf{2.863} & 2.592 & 0.910 & \underline{1.653} & 64.99 & 3.239 & 0.762 & \textbf{8.620} & \textbf{0.373} & 0.292 & \textemdash & \textemdash \\
& RMSE  & 4.352 & \textbf{0.475} & 3.556 &  \textbf{3.750} & 3.458 & 1.168 & \textbf{2.236} & 94.56 & 4.618 & 1.185 & \textbf{12.98} & \underline{0.719} & 0.450 & \textemdash & \textemdash \\
& nRMSD &  \textbf{0.284} & \textbf{0.259} & 0.296 &  \textbf{0.268} & \underline{0.282} & \underline{0.280} &  \textbf{0.318} & 0.368 & 0.310 & 0.341 & \textbf{0.305} & \textbf{0.288} & 0.278 & \underline{0.299} & \underline{0.298} \\
\midrule

CSDI
& MAE   & 3.794 & 0.411 & 2.916 & 3.528 & 2.828 & 0.931 & 2.016 & 65.59 & 3.692 & 0.816 & 9.693 & 0.437 & 31.65 & \textemdash & \textemdash \\
& RMSE  & 5.155 & 0.538 & 3.807 & 4.678 & 3.670 & 1.226 & 2.727 & 94.64 & 5.433 & 1.221 & 15.04 & 0.817 & 46.32 & \textemdash & \textemdash \\
& nRMSD & 0.355 & 0.289 & 0.342 & 0.352 & 0.313 & 0.306 & 0.465 & 0.406 & 0.397 & 0.442 & 0.381 & 0.391 & 0.312 & 0.366 & 0.365 \\
\midrule

TabPFN 
& MAE   & 3.510 & 0.461 & 2.991 & 3.128 & 2.830 & 0.956 & 2.149 & 75.80 & 3.359 & 0.975 & 10.23 & 0.583 & 32.16 & \textemdash & \textemdash \\
& RMSE  & 4.587 & 0.591 & 3.731 & 4.264 & 3.713 & 1.195 & 3.016 & 106.1 & 4.923 & 1.524 & 17.65 & 1.180 & 47.05 & \textemdash & \textemdash \\
& nRMSD & 0.330 & 0.311 & 0.335 & 0.312 & 0.318 & 0.302 & 0.544 & 0.517 & 0.345 & 0.525 & 0.391 & 0.467 & 0.308 & 0.387 & 0.385 \\
\midrule

\textbf{MA-PFN}
& MAE   & \underline{3.247} & \underline{0.384} & \textbf{2.557} & \underline{2.963} & \underline{2.553} & \underline{0.871} & 1.709 & \underline{64.85} & 3.200 & \underline{0.761} & \underline{8.930} & \textbf{0.373} & \textbf{28.38} & \textemdash & \textemdash \\
& RMSE  & \underline{4.293} & \underline{0.508} & \textbf{3.349} & \underline{3.963} & \underline{3.327} & \underline{1.127} & \underline{2.386} & 92.89 & \textbf{4.421} & \underline{1.163} & 13.37 & \textbf{0.702} & \textbf{43.72} & \textemdash & \textemdash \\
& nRMSD & \textbf{0.284} & \underline{0.267} & \textbf{0.286} & \underline{0.279} & \textbf{0.265} & \textbf{0.266} & \underline{0.333} & \underline{0.357} & \textbf{0.307} & \underline{0.338} & \underline{0.317} & \underline{0.296} & \textbf{0.271} & \textbf{0.297} & \textbf{0.297} \\
\bottomrule
\end{tabular}
\caption{Per-laboratory results for external forecasting on eICU. The first two weeks are used as context and observed laboratory measurements in the subsequent week are prediction targets. ``No. Queries'' denotes the number of evaluated observed query points for each laboratory, with Overall reporting the total number of queries across all 13 laboratories. MAE and RMSE are reported in the original laboratory units; lower values are better for all metrics. Best results are shown in \textbf{bold} and second-best results are \underline{underlined}.}
\label{tab:app_eicu_forecasting}
\end{table*}

%

%


\end{document}